\documentclass{article}
\usepackage{iclr2027_conference,times}

\usepackage{amsmath,amsfonts,bm}

\def\eqref#1{equation~\ref{#1}}

\def\1{\bm{1}}

\DeclareMathAlphabet{\mathsfit}{\encodingdefault}{\sfdefault}{m}{sl}
\SetMathAlphabet{\mathsfit}{bold}{\encodingdefault}{\sfdefault}{bx}{n}

\newif\ifinternaldraft
\internaldraftfalse

\newcommand{\method}{\textsc{GUI-HARVEST}\xspace}

\newcommand{\harnessspace}{\mathcal{H}}
\newcommand{\bundle}{\mathcal{B}}
\newcommand{\ledger}{\mathcal{L}}
\newcommand{\rollout}{\tau}

\newcommand{\searchset}{\mathcal{S}}
\newcommand{\valset}{\mathcal{V}}
\newcommand{\testset}{\mathcal{T}}

\usepackage{amsmath,amssymb}
\usepackage{booktabs}
\usepackage{tabularx}
\usepackage{makecell}
\usepackage{colortbl}
\usepackage{graphicx}
\usepackage{tikz}
\usepackage{microtype}
\usepackage{xcolor}
\usepackage[skins,breakable]{tcolorbox}
\usepackage{fancyvrb}
\usepackage{xspace}
\usepackage{multirow}
\usepackage{longtable}
\usepackage{capt-of}
\usepackage{pifont}
\usepackage{algorithm}
\usepackage{algpseudocode}
\usepackage{hyperref}
\hypersetup{hidelinks}
\usepackage{xurl}

\title{GUI-HARVEST: Self-Improving GUI Agents through Evidence-Driven Harness Evolution}

\author{Geyi Yang$^{1}$,
Zikun Qu$^{1}$,
Xiang Li$^{2}$,
Zhiyong Wang$^{3}$,
Min Zhang$^{4}$,\\
\textbf{Shipei Zeng}$^{5}$,
\textbf{Zhongxiang Dai}$^{1}$%
\thanks{Corresponding author. Correspondence to daizhongxiang@cuhk.edu.cn.}
\\[1ex]
$^{1}$The Chinese University of Hong Kong, Shenzhen \\
$^{2}$Tianjin University \quad
$^{3}$Harbin Institute of Technology (Shenzhen) \\
$^{4}$East China Normal University \\
$^{5}$Shenzhen Research Institute of Big Data
}

\iclrfinalcopy

\begin{document}
\raggedbottom

\setcounter{footnote}{1}
\maketitle

\begin{abstract}
The executable harness surrounding a GUI model determines how observations are
assembled, actions are executed, and verification, recovery, and termination
are controlled. Compared with harness optimization for non-GUI agents, automatically optimizing
this harness poses three coupled challenges: reconciling model intent with
observed visual effects, diagnosing failures under variable execution outcomes,
and identifying recurrent failure patterns across tasks and translating them
into reusable runtime changes.
We introduce \method, an automatic harness optimizer
that enables self-improving GUI agents with frozen backbone models.
First, to ground diagnosis in observed action effects, it aligns model outputs
and executed actions with before-and-after screenshots, producing findings
tied to specific interface transitions.
Second, to account for execution variability, it treats repeated executions
of the same task as a joint evidence unit, using within-task comparisons to
locate outcome-relevant behavioral differences.
Third, to derive reusable interventions from task-local evidence, it
consolidates verified findings across tasks into recurring failure patterns,
maps them to bounded source-code edits with predictions recorded before
evaluation, and checks the predicted behavioral effects alongside task
performance through repeated execution.
Experiments on OSWorld-Verified show consistent held-out gains 
across six general-purpose open, GUI-specialized open, and proprietary
backbone models;
Qwen3-VL-32B-Instruct gains 12.33 points on the
full suite.
Frozen-harness transfer improves GPT-5 by 13.87 percentage points on
WindowsAgentArena at 50 steps without further optimization.
With the same backbone and initial harness, \method
outperforms Self-Harness and Meta-Harness, suggesting that GUI-specific
diagnosis and validation help produce harness improvements that generalize
to unseen tasks.
The code is available at
\url{https://github.com/GaryYang12345/GUI-HARVEST}.

\end{abstract}

\section{Introduction}
\label{sec:introduction}
GUI agents rely on an executable \emph{runtime harness} to construct observations
and context, execute actions, and control verification, recovery, and
termination. Agent S and Agent S2/S3 demonstrate how these runtime choices
affect computer-use performance
\citep{agashe2024agents,agashe2025agents2,gonzalezpumariega2025scaling}.
Harness adaptation enables self-improving GUI agents whose own executions guide reusable changes to the runtime while model weights remain fixed.
Building
on methods that retain improvements from execution experience
\citep{shinn2023reflexion,zelikman2024stop}, automatic harness
optimizers such as Meta-Harness and Self-Harness revise prompts and code using
textual traces and feedback in non-GUI domains, including mathematical
reasoning and software engineering
\citep{lee2026metaharness,zhang2026selfharness}. GUI harness optimization poses
three coupled challenges that these generic methods do not directly address.

\textbf{First}, \emph{model intent must be reconciled with observed visual effects}.
A textual trace may claim success while the screen shows an unresolved state:
clicking Save, for example, does not complete the task if a dialog remains open.
\textbf{Second}, \emph{failures must be diagnosed under variable execution outcomes}.
The same agent can succeed or fail on repeated executions of the same task,
even under deterministic decoding
\mbox{\citep{gonzalezpumariega2026reliability}}. In our baseline runs,
11.8--20.4\% of Full tasks produce both zero and positive scores across
three executions, depending on the backbone
(Appendix Table~\ref{tab:fullset_flip}). A single failed trajectory is therefore
an incomplete account of the agent's behavior; other runs may reveal
successful alternatives and where execution diverged.
\textbf{Third}, \emph{task-specific findings must yield reusable harness changes}.
A patch motivated by one task need not help elsewhere. Similar failures
across tasks must be linked to a shared runtime mechanism and translated
into edits that generalize. Such edits must also be checked against the shared
failure pattern: higher aggregate scores alone do not show that the recurring
failure was corrected.

\begin{figure*}[t]
  \centering
  \vspace{-2mm}
  \includegraphics[width=\textwidth]{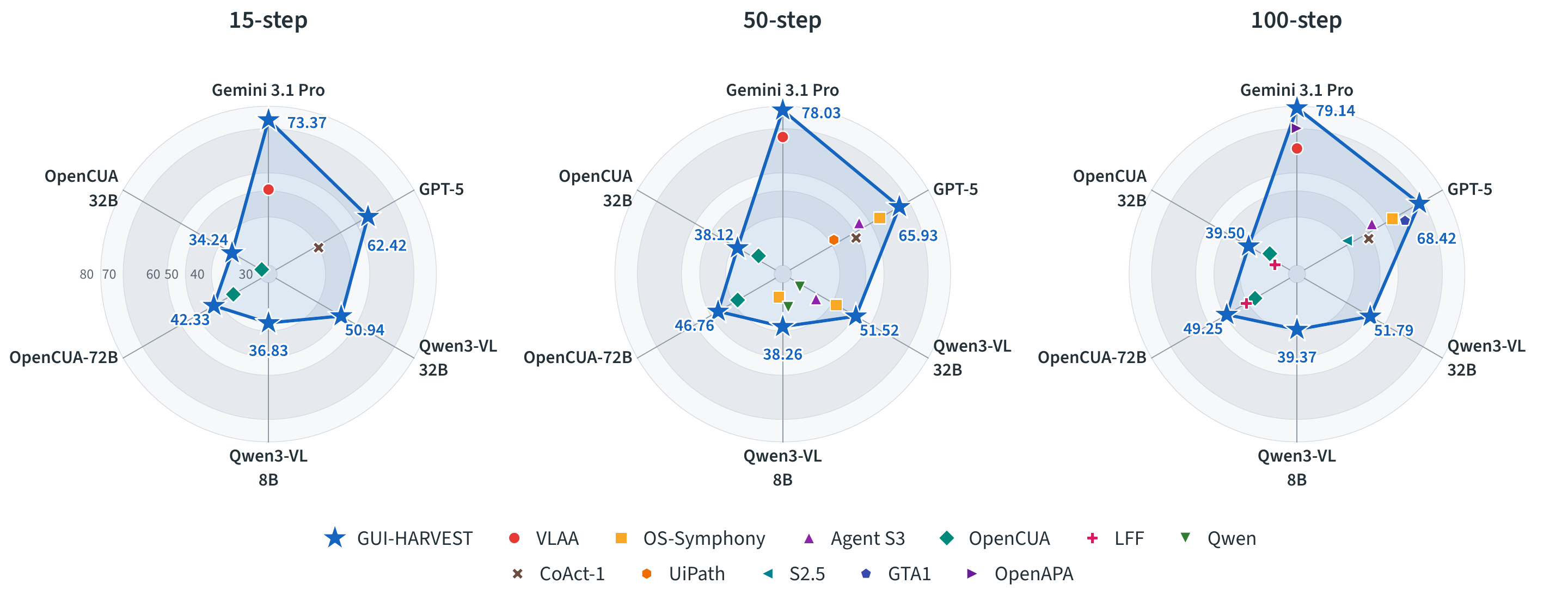}
  \vspace{-7mm}
  \caption{OSWorld-Verified results for six backbones at 15, 50, and 100 maximum
  environment steps. Axes fix the backbone; markers compare harnesses and agent
  frameworks at each budget.
  GUI-HARVEST uses only 15-step Search rollouts for optimization and remains
  frozen for evaluation, achieving the highest score in every model--budget
  comparison shown.}
  \label{fig:benchmark_radar}
\end{figure*}

\begin{figure*}[t]
    \centering
    \includegraphics[width=\textwidth]{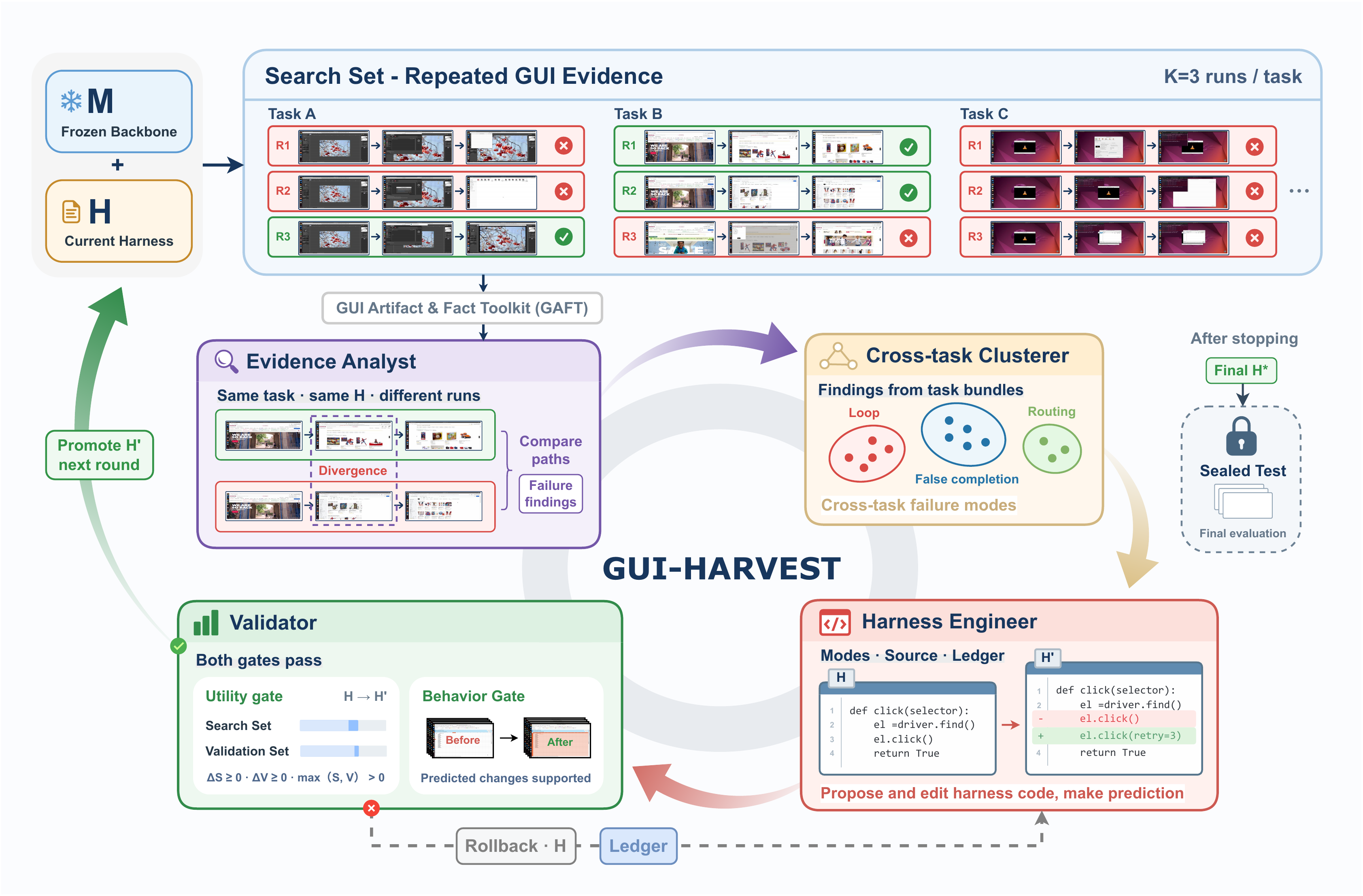}
    \caption{\method self-improvement loop. A frozen backbone and the current harness produce
    repeated search executions ($K=3$), which support
    task-local diagnosis, cross-task clustering, and source-code edits with
    explicit behavioral predictions. The Validator combines score and behavior
    checks to promote or roll back each edit; the test set remains sealed until
    optimization stops.}
    \label{fig:harvest_overview}
\end{figure*}

To address the challenges above, we introduce \method (\textbf{H}arness \textbf{A}daptation from
\textbf{R}epeated \textbf{V}isual \textbf{E}xecutions and
\textbf{S}tructured \textbf{T}rajectories), which automatically optimizes
harnesses for GUI agents with frozen backbones through three corresponding
design choices (Figure~\ref{fig:harvest_overview}).
\textbf{First}, to ground diagnosis in visual effects, the Evidence Analyst
aligns model outputs and executed actions with before-and-after screenshots.
The resulting findings identify discrepancies between intent and observed
state changes and retain links to supporting text and images.
\textbf{Second}, to account for execution variability, it compares repeated
executions of the same task as a joint evidence unit, using successful paths
when available to identify outcome-relevant behavioral differences. \textbf{Third}, to derive reusable interventions, the Cross-task Clusterer groups
verified findings into \emph{recurring failure patterns}. The Harness Engineer maps
these patterns to bounded source-code edits and records predictions of
observable behavioral effects of these edits before evaluation. The Validator promotes edits
only
when repeated executions pass score checks on search and held-out validation
tasks and behavioral checks of the predictions. An update ledger records
decisions to guide subsequent proposals. The test set remains sealed until
optimization ends.

We evaluate \method on OSWorld-Verified across six backbones spanning
general-purpose open, GUI-specialized open, and proprietary models. At 15 steps,
Test scores improve by 1.49--10.16 percentage points across all six
frozen backbones. Qwen3-VL-32B-Instruct achieves the largest full-suite gain,
reaching 50.94\% (+12.33 points; Table~\ref{tab:cross_backbone}). Harnesses
optimized at 15 steps improve further at 50- and 100-step evaluation budgets,
with Gemini 3.1 Pro reaching 79.14\% at 100 steps without further optimization
(Figure~\ref{fig:benchmark_radar}). Frozen-harness transfer to WindowsAgentArena (WAA)
at 50 steps raises Qwen3-VL-32B-Instruct and GPT-5 scores by 6.47 and 13.87
points, respectively, \emph{without WAA optimization} (Table~\ref{tab:waa}).
On Qwen3-VL-32B-Instruct, \method also outperforms
Self-Harness and Meta-Harness from the same initial harness, suggesting that
GUI-specific diagnosis and validation help produce harness improvements that
generalize better to unseen tasks. Its final harnesses also outperform LFF's
released harnesses on both OpenCUA backbones
(Table~\ref{tab:optimizer_comparison}).

Our contributions are threefold:
\begingroup
\setlength{\leftmargini}{1.5em}
\setlength{\topsep}{2pt}
\setlength{\partopsep}{0pt}
\setlength{\itemsep}{1pt}
\setlength{\parsep}{0pt}
\begin{itemize}
    \item \textbf{Evidence-driven self-improvement for GUI agents.} We formulate harness
    adaptation as evidence-driven improvement of an executable runtime around
    a frozen GUI model.
    \item \textbf{Multimodal diagnosis and intervention validation.}
    We propose \method to convert repeated multimodal execution evidence into
    reusable code changes validated by score and behavior checks.
    \item \textbf{Evaluation across backbones and environments.}
    We demonstrate Test gains across six backbones, performance at larger
    step budgets, and frozen-harness transfer beyond OSWorld.
\end{itemize}
\endgroup

\section{Problem setting}
\label{sec:formulation}

Let $M$ be a frozen GUI-capable model and $H_0\in\harnessspace$ an executable
harness. A harness contains the editable program around the model: prompts and
context construction, memory, action and tool interfaces, control flow,
verification, recovery, and termination. The optimizer cannot modify model
weights, benchmark tasks, evaluators, environment infrastructure, or resource
limits. A
capability manifest specifies which harness files may be edited and where new
modules may be created.

Before optimization, tasks are divided into search $\searchset$, validation
$\valset$, and sealed test $\testset$ by a domain-stratified random split with
fixed per-domain proportions. Search trajectories supply optimization evidence.
Validation trajectories are hidden from the optimizer and contribute
only aggregate scores. The sealed test set is evaluated after optimization
stops. An optimization round is the search performed at a fixed harness state
$H_t$. The optimizer may evaluate up to $P$ L1-valid candidate edits in that
round. Only promotion sets $H_{t+1}$ and starts the next round; exhausting all
$P$ attempts terminates optimization at $H_t$. We cap the process at $B$
rounds. The optimizer returns
$\mathcal{O}(M,H_0,\searchset,\valset,B,P)\rightarrow H^*$.
For a task $x_i$, one run under harness $H$ is
$\rollout_{i,r}(H)=\rollout(M,H,x_i;\xi_{i,r})$,
where $\xi_{i,r}$ includes model sampling, rendered observations, interface
timing, and application responses. If $R(\rollout)\in[0,1]$ is the benchmark's
raw evaluator score, the repeated mean on a task set $D$ is
\begin{equation}
\label{eq:utility}
    \widehat J_{D,K}(H)=\frac{1}{|D|}\sum_{i\in D}
    \frac{1}{K}\sum_{r=1}^{K}R\!\left(\rollout_{i,r}(H)\right).
\end{equation}

\section{GUI-HARVEST}
\label{sec:method}
In this paper, a \emph{task} is a user instruction to achieve a specified outcome in one or
more desktop applications, together with an initial environment state and an
evaluator that scores the resulting state.
\textcolor{black}{\method improves the executable \emph{harness} around a frozen
\emph{backbone}.} As shown in
Figure~\ref{fig:harvest_overview}, repeated task executions supply evidence for
four optimizer roles: the Evidence Analyst, Cross-task Clusterer, Harness
Engineer, and Validator. Together, they connect observed failures to reusable
code changes and test whether those changes have their intended effects.


\subsection{The Self-Improvement Loop}
\label{sec:iteration}

\textcolor{black}{Starting from $H_0$, we collect $K$ \emph{rollouts} (complete task
executions) per search and validation task, reusing runs already available for
the current harness.} Search provides detailed execution evidence;
validation exposes only aggregate scores. Each round diagnoses failures under
$H_t$, groups recurring behaviors, and proposes and evaluates an edited harness
$\widetilde H_t$. \emph{Promotion} adopts a candidate that passes the performance
and behavior checks; \emph{rollback} restores $H_t$ after rejection. An
\emph{update ledger}, a record of edits, predictions, scores, and decisions,
guides subsequent proposals.
This closes a self-improvement loop in which accepted harness
changes reshape subsequent executions, supplying evidence for further adaptation.

After promotion, we analyze the candidate's search rollouts and rebuild the
behavior modes. After rejection, we retain the current harness's verified
findings and modes, and the engineer proposes a materially different edit.
\textcolor{black}{Appendix Algorithm~\ref{alg:harvest}} summarizes this loop; Appendix~\ref{app:implementation}
specifies the role interfaces, and Appendix~\ref{app:formal_method} gives the
formal record definitions and promotion rule.

A candidate that passes the L0/L1 code checks (Section~\ref{sec:validation}) and
enters GUI evaluation consumes one attempt in the current round. Promotion
advances to the next round and opens a fresh attempt budget; if all $P=5$
attempts at a harness state are rejected, optimization stops. The main protocol
permits at most $B=10$ rounds. L0/L1 repairs made before a valid candidate
enters GUI evaluation consume neither budget. \textcolor{black}{After stopping, we freeze $H^*$ for sealed-test
evaluation; the test set remains unavailable during optimization.}

\subsection{Evidence Analyst: Diagnosis from Repeated Executions}
\label{sec:evidence}

The Evidence Analyst compares repeated runs of the same task to produce
\emph{findings}: evidence-supported descriptions of problematic behavior in a
specific run. This \emph{task-local} analysis keeps the task and harness fixed;
successful runs, when available, provide alternative paths to completion.

\textbf{Task-round bundles.}
Each task is run $K$ times ($K=3$ in the main setting) from
\emph{clean environment snapshots} that restore its initial state.
\textcolor{black}{Runs share the model, harness behavior hash, action limit, and
decoding settings.} Each run records a
\emph{multimodal trajectory}---text, planned and executed actions, screenshots,
and tool outputs at each step---along with its benchmark score and runtime and
termination metadata. \textcolor{black}{A \emph{task-round bundle} groups these
runs with a Task Card containing the instruction, domain, related applications,
feasibility annotation, setup steps, and any published hint, together with a
\emph{Harness Card} summarizing the runtime and its observable outputs.}

A bundle enters diagnosis when at least one valid run scores zero. The analyst
diagnoses these zero-score runs, using the remaining runs as comparison
evidence. All valid scores, including partial scores, contribute to
Equation~\ref{eq:utility}.

\textbf{Organizing execution evidence.}
We use the GUI Artifact and Fact Toolkit (GAFT) as supporting infrastructure to
align text, actions, and before-and-after screenshots by execution step,
retaining links to tool outputs, final scores, and evidence files. It computes
\emph{mechanical facts}, such as repeated actions and pixel changes; the analyst
interprets their relevance to failure. \textcolor{black}{A visual index supports
retrieval of complete traces and full-resolution screenshots.}

\textbf{Diagnosing failures.}
One budget-limited analyst examines each bundle independently. It compares the
\textcolor{black}{\emph{terminal application state}} with the task
objective, then locates where runs diverge and checks whether intended actions,
executed actions, and subsequent screens agree. \textcolor{black}{A finding
identifies a decisive step in a failed run, describes the observed behavior and
machine outcome, and links the diagnosis to run/step evidence.}
A \emph{deterministic verifier} checks
the harness version, cited run and step, quotation locality, and mechanically
testable trajectory facts before retaining the finding. These checks establish
evidence consistency, while the behavioral interpretation remains the
analyst's assessment.

\subsection{Cross-task Clusterer: Recurring Behavior Modes}
\label{sec:analysis}

The Cross-task Clusterer groups verified findings from independent tasks into
\emph{behavior modes}: recurring patterns that suggest shared harness changes.
To keep this induction grounded without imposing a fixed mode taxonomy, each
finding is first assigned a deterministic outcome category from its decisive
action and observed machine result. Within each category, the clusterer uses
the finding's mechanism and evidence-supported counterfactual to discover
open-vocabulary modes shared by at least two tasks. The categories constrain
the comparison space; they do not prescribe mode names or code patches.

Each mode records an observable mechanism, a membership test, a behavioral
target, its member findings, and independent-task support. Member identifiers
retain the link to task-local evidence. \textcolor{black}{Modes are stored as
versioned round artifacts and regenerated from the updated harness's executions
after promotion.}

\subsection{Harness Engineer: Source-Aware Interventions}
\label{sec:engineering}

\textcolor{black}{The Harness Engineer translates behavior modes into harness
code changes.} Its \emph{source-aware} session allows code
inspection, editing, and testing using the modes, representative evidence,
Harness Card, and relevant rejected attempts from the ledger.
\textcolor{black}{Within the capability manifest, the engineer selects a small
compatible set of modes, locates the relevant runtime code, applies a bounded
\emph{patch}, and runs local tests.}

Before GUI evaluation, the engineer records an update plan linking the target
modes and evidence to the \emph{source surface} (code region), proposed edit,
observable predictions, and search tasks used to check them. The plan also
specifies \emph{invariants}: interfaces or behaviors that must remain intact.
Recording predictions before evaluation
makes the edit testable: a completion-check patch might predict that the agent
resolves an open save dialog before declaring success.

\subsection{Validator: Utility and Behavioral Checks}
\label{sec:validation}

The Validator checks task performance and predicted behavioral changes.
\textcolor{black}{Two deterministic code checks first screen $\widetilde H_t$:
L0 checks edit permissions, diff limits, syntax, and static rules; L1 checks
imports, interface compatibility, unit tests, and a \emph{smoke test}.}
Passing candidates are run $K$ times on every search and validation task under
the fixed execution-cost constraint. This batch supplies scores and post-edit
bundles for the checks below; promoted candidates also supply the next round's
search evidence.

\begin{samepage}
\textbf{Utility hard gate.}
\textcolor{black}{Using the utility in Equation~\ref{eq:utility}, let} $\Delta_D$ denote the
candidate's mean score minus the current harness's mean score on split
$D\in\{\searchset,\valset\}$.
The hard gate requires no decline on either split and a strict gain on at least
one:
\begin{equation}
\label{eq:hardgate}
    \operatorname{HardGate}_{\searchset,\valset}=
    (\Delta_{\searchset}\geq0)\land(\Delta_{\valset}\geq0)
    \land\bigl(\max(\Delta_{\searchset},\Delta_{\valset})>0\bigr).
\end{equation}
Both splits contribute scores, but only search executions are available for
diagnosis and behavioral validation.
\end{samepage}

\textbf{Behavioral soft gate.}
This required check tests whether behavior changed as predicted. For each
targeted search task, a separate validator compares the frozen prediction and
pre-edit task-local finding with the post-edit $K$-run trajectories and
available screen evidence. It
returns \emph{supported} if the evidence agrees with the prediction,
\emph{contradicted} if it conflicts, or \emph{inconclusive} if insufficient,
citing verifiable runs and steps. A deterministic aggregator applies a fixed
rule across task verdicts: each prediction needs at least one \emph{determinate}
verdict (supported or contradicted) and more supporting than contradicting
verdicts. Each validator's visual context is restricted to one task bundle.

The candidate is promoted exactly when it passes L0, L1, the utility hard gate,
and the behavioral soft gate. Promotion adopts the candidate, updates the
Harness Card, and retains new search bundles. Rejection restores $H_t$ and its
verified evidence and modes.
The ledger records the patch, predictions, score changes, and verdicts.
\textcolor{black}{The loop repeats until a stopping condition in
Section~\ref{sec:iteration} is met.}


\section{\textcolor{black}{Experiments}}
\label{sec:results}

\begingroup
\color{black}
\subsection{Experimental setup}
\label{sec:experiments}

\textbf{Benchmarks and metrics.}
Following common OSWorld evaluation practice, we use OSWorld-Verified with 361
tasks after excluding eight Google Drive tasks
\citep{xie2024osworld,han2026vlaa}. A fixed, domain-stratified random split
assigns 80 tasks to Search, 80 to Validation, and 201 to sealed Test while
preserving each domain's proportion. Only Search trajectories supply optimization evidence; Validation
exposes aggregate scores, and Test is used only after harness optimization. We
report mean raw evaluator scores (Eq.~\ref{eq:utility}) on each split and the
full 361-task suite, comparing the initial harness $H_0$ with the selected
harness $H^*$.

\textbf{Backbones and harnesses.}
We evaluate Qwen3-VL-8B/32B-Instruct, OpenCUA-32B/72B, Gemini 3.1 Pro, and
GPT-5, with target-model weights and decoding settings frozen
\citep{qwen2025qwen3vl,wang2025opencua}. Qwen3-VL and proprietary backbones
start from Agent S3 with its default grounding model (UI-TARS-1.5-7B) and
behavior best-of-$N$ disabled ($N=1$) \citep{gonzalezpumariega2025scaling}; OpenCUA models use their
official coordinate-action runtime. All four optimizer roles in our algorithm use Claude
Sonnet 5 with the same role prompts across backbones.

\textbf{Optimization and evaluation.}
Harness search uses a 15-step limit and $K=3$ independent runs per task. It
allows at most $B=10$ rounds, with up to $P=5$ evaluated candidate
edits per round; optimization stops when all five attempts at one harness state
fail promotion. Selected harnesses are frozen for
OSWorld evaluation at 15, 50, and 100 steps. Qwen3-VL-32B-Instruct and GPT-5
also transfer to WindowsAgentArena at 50 steps without WAA optimization
\citep{bonatti2025windows}. Ablations on Qwen3-VL-32B-Instruct remove visual
evidence, cross-task clustering, repeated optimizer evidence, or behavioral
validation, with all selected harnesses evaluated using $K=3$. Published
comparisons retain their reported runtimes and budgets.
Appendix~\ref{app:implementation} provides more implementation details and
comparison protocols.
\par
\endgroup

\subsection{Harness Adaptation and Cross-Benchmark Transfer}

\begin{table}[t]
\centering
\begin{minipage}[t]{0.48\textwidth}
\centering
\captionof{table}{Paired 15-step results on Test and Full.
Parentheses show gains over the corresponding initial harness.\vphantom{g}}
\label{tab:cross_backbone}
\vspace{2pt}
\tiny
\setlength{\tabcolsep}{1.6pt}
\renewcommand{\arraystretch}{1.31}
\definecolor{harvestTestBg}{HTML}{EEF3F8}
\definecolor{harvestFullBg}{HTML}{F7F1E5}
\begin{tabular*}{\linewidth}{@{\extracolsep{\fill}}l
>{\columncolor{harvestTestBg}}r
>{\columncolor{harvestFullBg}}r
>{\columncolor{harvestTestBg}}r
>{\columncolor{harvestFullBg}[\tabcolsep][0pt]}r@{}}
\toprule
& \multicolumn{2}{c}{Initial $H_0$} & \multicolumn{2}{c}{\method $H^*$} \\
\cmidrule(lr){2-3}\cmidrule(lr){4-5}
Backbone & Test & Full & Test & Full \\
\midrule
Qwen3-VL-8B & 28.79 & 29.49 & \textbf{34.84} {\color{green!50!black}(+6.05)} & \textbf{36.83} {\color{green!50!black}(+7.34)} \\
Qwen3-VL-32B & 43.02 & 38.61 & \textbf{53.18} {\color{green!50!black}(+10.16)} & \textbf{50.94} {\color{green!50!black}(+12.33)} \\
OpenCUA-32B & 32.82 & 30.71 & \textbf{34.31} {\color{green!50!black}(+1.49)} & \textbf{34.24} {\color{green!50!black}(+3.53)} \\
OpenCUA-72B & 41.25 & 37.65 & \textbf{44.18} {\color{green!50!black}(+2.93)} & \textbf{42.33} {\color{green!50!black}(+4.68)} \\
Gemini 3.1 Pro & 66.42 & 66.46 & \textbf{72.63} {\color{green!50!black}(+6.20)} & \textbf{73.37} {\color{green!50!black}(+6.91)} \\
GPT-5 & 56.35 & 55.20 & \textbf{61.86} {\color{green!50!black}(+5.51)} & \textbf{62.42} {\color{green!50!black}(+7.22)} \\
\bottomrule
\end{tabular*}
\end{minipage}\hfill
\begin{minipage}[t]{0.49\textwidth}
\centering
\captionof{table}{Harness-optimizer comparison at 15 steps. 
Experimental protocols are detailed in
Appendix~\ref{app:optimizer_comparison}.}
\label{tab:optimizer_comparison}
\vspace{2pt}
\tiny
\setlength{\tabcolsep}{1.8pt}
\renewcommand{\arraystretch}{1.08}
\begin{tabular*}{\linewidth}{@{\extracolsep{\fill}}lrr@{}}
\toprule
\rowcolor{black!7}\multicolumn{3}{@{}l}{\textit{General harness optimizers (GUI-adapted search)}} \\
Method & Test & Full \\
\midrule
Self-Harness & 45.02 {\color{green!50!black}(+2.00)} & 42.08 {\color{green!50!black}(+3.47)} \\
Meta-Harness & 47.40 {\color{green!50!black}(+4.38)} & 45.10 {\color{green!50!black}(+6.49)} \\
\textbf{\method} & \textbf{53.18} {\color{green!50!black}(+10.16)} & \textbf{50.94} {\color{green!50!black}(+12.33)} \\
\midrule
\rowcolor{black!7}\multicolumn{3}{@{}l}{\textit{GUI failure-driven optimizers (final harnesses)}} \\
Backbone & LFF & \method \\
\midrule
OpenCUA-32B & 31.83 / 32.12 & \textbf{34.31 / 34.24} \\
OpenCUA-72B & 41.29 / 39.50 & \textbf{44.18 / 42.33} \\
\bottomrule
\end{tabular*}
\end{minipage}
\end{table}

\begin{figure}[t]
  \centering
  \includegraphics[width=\textwidth]{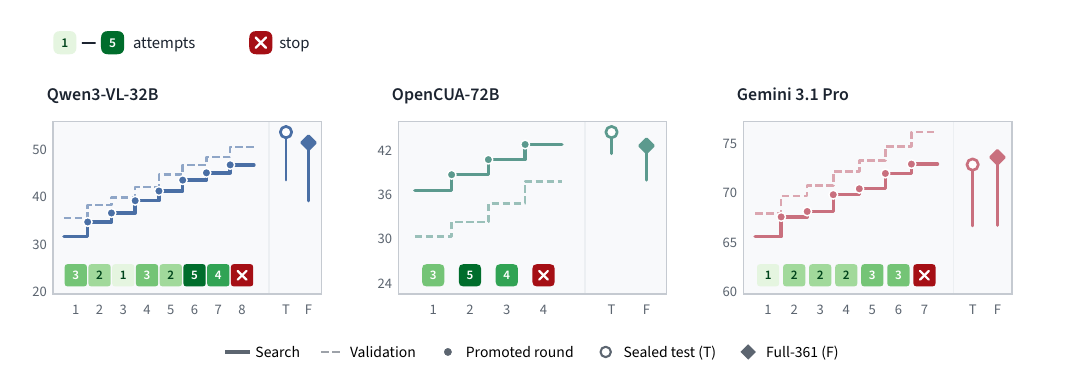}
  \caption{\textcolor{black}{Harness evolution for Qwen3-VL-32B, OpenCUA-72B,
  and Gemini 3.1 Pro. Solid and dashed curves show Search and Validation
  scores, respectively; filled circles mark promoted harnesses, and the
  rightmost markers show Test (T) and Full (F) scores after
  optimization stops. The numbered tiles report candidate-edit attempts in
  each optimization round, and the red cross marks the terminal round.
  Results for the other three backbones are plotted in Appendix
  Figure~\ref{fig:harness_evolutions}.}}
  \label{fig:harness_evolutions_main}
\end{figure}

\textbf{Harness adaptation across backbones.}
Under the unified 15-step, $K=3$ protocol, \method improves Test and Full
scores for all six frozen backbones (Table~\ref{tab:cross_backbone}).
Qwen3-VL-32B-Instruct achieves the largest gains, reaching 53.18\% on
Test (+10.16 percentage points) and 50.94\% on Full (+12.33 points).
These held-out gains show that harness adaptation generalizes beyond the
optimization tasks across both general-purpose and GUI-specialized backbones.
Appendix Table~\ref{tab:cross_backbone_full} provides complete
split-level results. 
Figure~\ref{fig:harness_evolutions_main}
traces the self-improvement process, showing Search and Validation gains
accumulating over successive promoted edits and variation in adaptation
effort across backbones.


\textbf{Comparison with harness optimizers.}
Table~\ref{tab:optimizer_comparison} shows that, from the same initial
Qwen3-VL-32B-Instruct/Agent S3 harness, \textbf{\method gains 10.16 Test points,
versus 2.00 for Self-Harness and 4.38 for Meta-Harness}
\citep{zhang2026selfharness,lee2026metaharness}. With the optimizer model and
ten-round cap also shared (Appendix~\ref{app:optimizer_comparison}), this
advantage suggests that GUI-specific diagnosis and validation yield more
generalizable edits than the evaluated generic search procedures.
\method also outperforms LFF~\citep{sun2026learning} on both OpenCUA backbones;
on OpenCUA-72B, Test and Full scores for \method reach 44.18\% and 42.33\%,
versus 41.29\% and 39.50\% for LFF. 
For LFF, we evaluate the released intervention patches under the same protocol, as the optimizer implementation and complete search configuration are not publicly available; this comparison therefore focuses on the performance of the resulting harnesses.

\textbf{Transfer to WindowsAgentArena.}
We additionally transfer the frozen Qwen3-VL-32B-Instruct and GPT-5 harnesses
to WAA at a 50-step budget, without using WAA trajectories for optimization.
Qwen3-VL-32B-Instruct improves from 38.21\% to 44.68\% (+6.47), while GPT-5
improves from 50.88\% to 64.75\% (+13.87). With only platform adapters changed,
these gains suggest that runtime improvements learned on OSWorld remain useful
under a different platform and task distribution. Appendix
Tables~\ref{tab:waa} and~\ref{tab:waa_benchmark} report the paired transfer
results and six-domain benchmark context, respectively.



\subsection{Matched-Backbone OSWorld Comparison}

Figure~\ref{fig:benchmark_radar} compares the harnesses optimized by \method
against existing widely adopted GUI models and harnesses, with
backbones and step budgets matched. The selected harnesses achieve \textbf{the highest
score in every comparison shown, spanning six backbones and 15-, 50-, and
100-step budgets.} These advantages extend across general open, GUI-specialized,
and proprietary backbones and persist at larger evaluation budgets: all
\method harnesses are optimized at 15 steps and then frozen.

The largest gains include \textbf{22.61 percentage points over CoAct-1 with GPT-5}
(62.42\% versus 39.81\%) and \textbf{21.68 points over VLAA-GUI with Gemini 3.1 Pro}
(73.37\% versus 51.69\%) at 15 steps. At 50 steps, the selected
Qwen3-VL-32B-Instruct harness reaches 51.52\%, exceeding the Qwen baseline
(32.60\%) by 18.92 points. Figure~\ref{fig:benchmark_bars} provides more
detailed comparisons, and Appendix~\ref{app:additional} reports domain-level
results and source details.

%
%

\textcolor{black}{Adaptation gains depend on more than model capability:
Qwen and frontier backbones gain 6.91--12.33 Full points over their initial
harnesses, versus 3.53--4.68 for OpenCUA (Table~\ref{tab:cross_backbone}).
OpenCUA is post-trained around its native coordinate-bearing action protocol;
its smaller gains are consistent with lower adaptability to added tools and
control mechanisms.
Increasing the budget from 15 to 100 steps, however, adds only 2.54 and 0.85
points for Qwen3-VL-8B and 32B, versus 5.77 and 6.00 for Gemini 3.1 Pro and
GPT-5 (Figure~\ref{fig:benchmark_radar}). Frontier models thus benefit more
from additional interaction, while Qwen's limited gains suggest persistent
long-horizon execution bottlenecks
(Appendix Table~\ref{tab:profiles}).}

\subsection{Accuracy--Cost Operating Points}



\begin{figure}[t]
  \centering
  \vspace{-2mm}
  \includegraphics[width=0.9\textwidth]{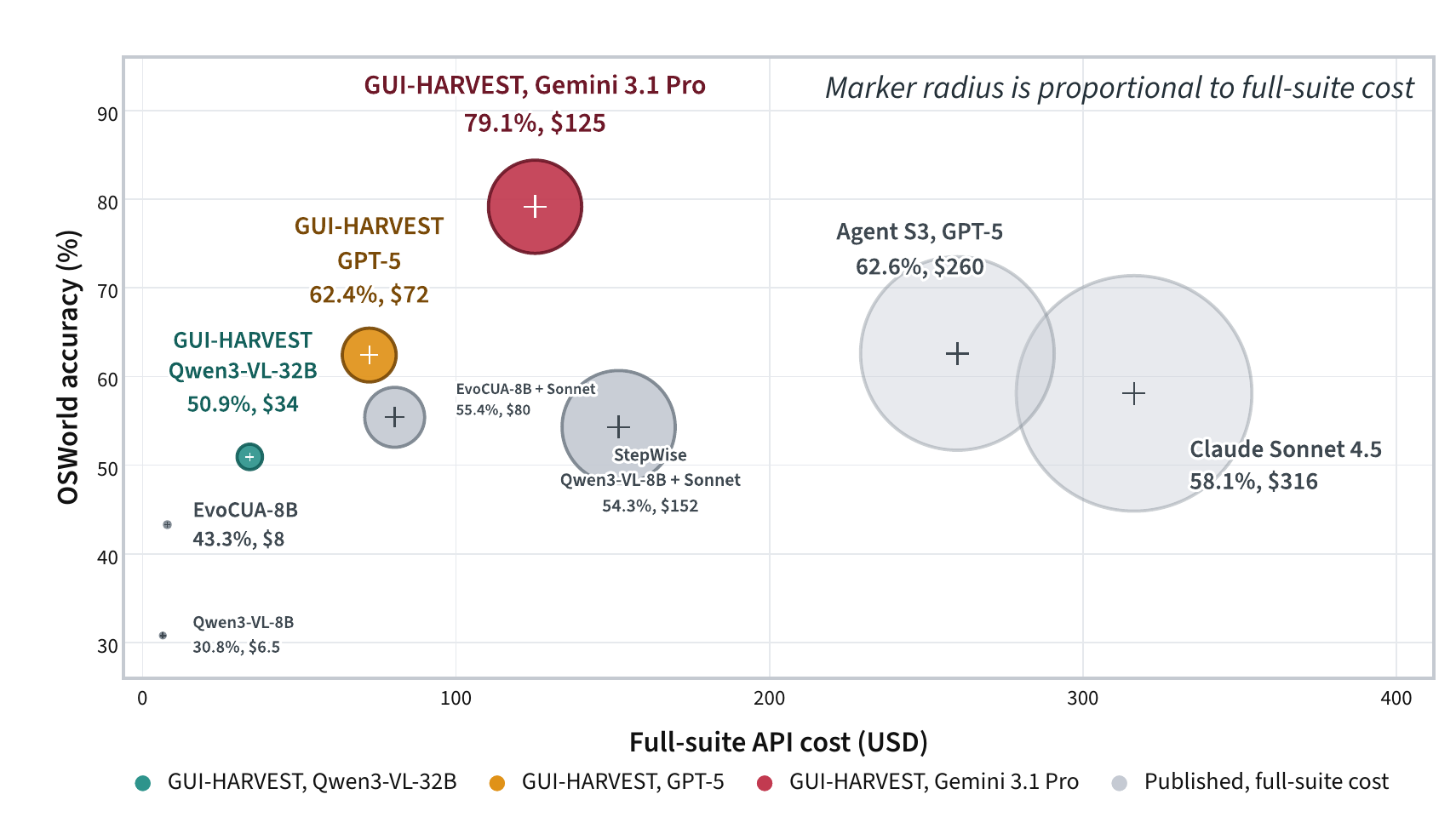}
    \vspace{-2mm}
  \caption{OSWorld accuracy versus full-suite API cost for systems with
  available cost records. We show the 15-step GUI-HARVEST points for
  Qwen3-VL-32B-Instruct and GPT-5 and the 100-step point for Gemini 3.1 Pro.
  Published points retain the models, harnesses, budgets, protocols, and
  per-task costs reported by
  \citet{wei2026steplevel,gonzalezpumariega2025scaling}; full-suite costs are
  reconstructed as detailed in Appendix~\ref{app:cost}.}
  \label{fig:benchmark_cost}
\end{figure}

Figure~\ref{fig:benchmark_cost} compares source-reported operating points, not
a shared-budget experiment: our GPT-5 point uses 15 steps, Agent S3/GPT-5 uses
100, and the other published points retain their source protocols.
With GPT-5, our 15-step harness \textbf{nearly matches Agent S3's accuracy
(62.4\% versus 62.6\%) at approximately 72\% lower full-suite API cost}
(\$72 versus \$260). With Gemini 3.1 Pro, our 100-step harness reaches
79.1\% at \$125, \textbf{exceeding Claude Sonnet 4.5 (58.1\% at \$316) by 21.0
percentage points while costing approximately 60\% less.}
Qwen3-VL-32B-Instruct offers a lower-cost operating point of 50.9\% at \$34.
A plausible explanation for our cost reductions is that the optimized harness reduces repeated failed
actions and unnecessary continuation, which both consume API calls and impede
task completion. Correcting these behaviors can therefore improve accuracy
and inference efficiency together. More detailed results are reported in Appendix~\ref{app:cost}.

\subsection{Component Ablations}




\begingroup
\definecolor{traceGreenOne}{HTML}{E5F5E0}
\definecolor{traceGreenTwo}{HTML}{A1D99B}
\definecolor{traceGreenThree}{HTML}{74C476}
\definecolor{traceGreenFour}{HTML}{31A354}
\definecolor{traceGreenFive}{HTML}{006D2C}
\definecolor{traceGreenText}{HTML}{00441B}
\definecolor{traceStop}{HTML}{A50F15}
\definecolor{traceUnused}{HTML}{F2F3F4}
\newcommand{\tracegap}{\hspace{0.34mm}}
\newcommand{\traceunit}[3]{%
  \tikz[baseline=(trace.base)]{\node[
    fill=#1, draw=none, rounded corners=1.25pt,
    minimum width=0.32cm, minimum height=0.32cm,
    inner sep=0pt, text=#2,
    font=\fontsize{4.4}{4.6}\selectfont\bfseries] (trace) {#3};}}
\newcommand{\traceone}{\traceunit{traceGreenOne}{traceGreenText}{1}}
\newcommand{\tracetwo}{\traceunit{traceGreenTwo}{traceGreenText}{2}}
\newcommand{\tracethree}{\traceunit{traceGreenThree}{white}{3}}
\newcommand{\tracefour}{\traceunit{traceGreenFour}{white}{4}}
\newcommand{\tracefive}{\traceunit{traceGreenFive}{white}{5}}
\newcommand{\tracenone}{\traceunit{traceUnused}{traceUnused}{\phantom{1}}}
\newcommand{\tracestop}{%
  \tikz[baseline=(trace.base)]{%
    \node[fill=traceStop, draw=none, rounded corners=1.25pt,
      minimum width=0.32cm, minimum height=0.32cm, inner sep=0pt,
      text=white, font=\fontsize{4.4}{4.6}\selectfont\bfseries]
      (trace) {\phantom{5}};
    \draw[white, line width=0.90pt, line cap=round]
      ([xshift=3.00pt,yshift=3.00pt]trace.south west) --
      ([xshift=-3.00pt,yshift=-3.00pt]trace.north east);
    \draw[white, line width=0.90pt, line cap=round]
      ([xshift=3.00pt,yshift=-3.00pt]trace.north west) --
      ([xshift=-3.00pt,yshift=3.00pt]trace.south east);}}
\newcommand{\tracelegunit}[3]{%
  \tikz[baseline=(trace.base)]{\node[
    fill=#1, draw=none, rounded corners=1pt,
    minimum width=0.24cm, minimum height=0.24cm,
    inner sep=0pt, text=#2,
    font=\fontsize{3.8}{4.0}\selectfont\bfseries] (trace) {#3};}}
\newcommand{\tracelegone}{\tracelegunit{traceGreenOne}{traceGreenText}{1}}
\newcommand{\tracelegfive}{\tracelegunit{traceGreenFive}{white}{5}}
\newcommand{\tracelegstop}{%
  \tikz[baseline=(trace.base)]{%
    \node[fill=traceStop, draw=none, rounded corners=1pt,
      minimum width=0.24cm, minimum height=0.24cm, inner sep=0pt,
      text=white, font=\fontsize{3.8}{4.0}\selectfont\bfseries]
      (trace) {\phantom{5}};
    \draw[white, line width=0.68pt, line cap=round]
      ([xshift=2.20pt,yshift=2.20pt]trace.south west) --
      ([xshift=-2.20pt,yshift=-2.20pt]trace.north east);
    \draw[white, line width=0.68pt, line cap=round]
      ([xshift=2.20pt,yshift=-2.20pt]trace.north west) --
      ([xshift=-2.20pt,yshift=2.20pt]trace.south east);}}
\newcommand{\tracelegnone}{\tracelegunit{traceUnused}{traceUnused}{\phantom{1}}}
\newcommand{\tracehead}[1]{\makebox[0.32cm][c]{\tiny #1}}
\newcommand{\tracelegend}{%
  \makecell[c]{\textbf{Optimization trace}\\[-1.5pt]
  \tracelegone\,--\,\tracelegfive\;\tiny attempts\quad
  \tracelegstop\;\tiny stop\quad
  \tracelegnone\;\tiny unused}}
\newcommand{\metriccell}[2]{%
  #1\,{\color{green!50!black}(+#2)}}
\newcommand{\metriccellbf}[2]{%
  \textbf{#1}\,{\color{green!50!black}\textbf{(+#2)}}}

\begin{table}[t]
\vspace{-3mm}
\centering
\caption{Component ablations and optimization traces on
Qwen3-VL-32B-Instruct. Test and Full gains are measured against the
initial harness. Each numbered trace cell denotes a promoted round; its value
and shade encode the number of edit attempts required for promotion. X marks
termination. All selected harnesses use the same $K=3$ final evaluation protocol.}
\label{tab:ablation_joint}
\scriptsize
\renewcommand{\arraystretch}{1.24}
\setlength{\tabcolsep}{1.5pt}
\begin{tabular*}{\textwidth}{@{\extracolsep{\fill}}p{0.30\textwidth}cccc@{}}
\toprule
\multicolumn{4}{c}{\textbf{Final harness}} & \tracelegend \\
\cmidrule(lr){1-4}\cmidrule(l){5-5}
\textbf{Variant} & \textbf{Test} & \textbf{Full} & \textbf{Rounds} &
\tracehead{1}\tracegap\tracehead{2}\tracegap\tracehead{3}\tracegap
\tracehead{4}\tracegap\tracehead{5}\tracegap\tracehead{6}\tracegap
\tracehead{7}\tracegap\tracehead{8}\tracegap\tracehead{9}\tracegap
\tracehead{10} \\
\midrule
w/o visual evidence
& \metriccell{44.54}{1.52}
& \metriccell{40.69}{2.08}
& 4
& \tracefour\tracegap\tracefour\tracegap\tracefive\tracegap\tracestop\tracegap
  \tracenone\tracegap\tracenone\tracegap\tracenone\tracegap\tracenone\tracegap
  \tracenone\tracegap\tracenone \\
w/o Cross-task Clusterer
& \metriccell{45.66}{2.64}
& \metriccell{42.13}{3.52}
& 4
& \tracefour\tracegap\tracefive\tracegap\tracefour\tracegap\tracestop\tracegap
  \tracenone\tracegap\tracenone\tracegap\tracenone\tracegap\tracenone\tracegap
  \tracenone\tracegap\tracenone \\
Single-run optimizer evidence ($K=1$)
& \metriccell{47.44}{4.42}
& \metriccell{43.12}{4.51}
& 5
& \tracethree\tracegap\tracefive\tracegap\tracetwo\tracegap
  \tracefour\tracegap\tracestop\tracegap
  \tracenone\tracegap\tracenone\tracegap\tracenone\tracegap\tracenone\tracegap\tracenone \\
Hard gate only
& \metriccell{49.36}{6.34}
& \metriccell{48.08}{9.47}
& 9
& \tracetwo\tracegap\traceone\tracegap\tracetwo\tracegap\traceone\tracegap
  \tracetwo\tracegap\tracethree\tracegap\traceone\tracegap\tracefour\tracegap
  \tracestop\tracegap\tracenone \\
\textbf{Full \method}
& \metriccellbf{53.18}{10.16}
& \metriccellbf{50.94}{12.33}
& \textbf{8}
& \tracethree\tracegap\tracetwo\tracegap\traceone\tracegap\tracethree\tracegap
  \tracetwo\tracegap\tracefive\tracegap\tracefour\tracegap\tracestop\tracegap
  \tracenone\tracegap\tracenone \\
\bottomrule
\end{tabular*}
\end{table}
\endgroup

All four ablations underperform the complete optimization loop
(Table~\ref{tab:ablation_joint}), supporting the usefulness of each component.
Full \method reaches 53.18\% on Test and 50.94\% on Full.
Removing visual evidence produces the largest drop, to 44.54\% and 40.69\%,
highlighting its value for GUI failure diagnosis. Removing the Cross-task
Clusterer yields 45.66\% and 42.13\%, supporting cross-task aggregation to
guide reusable harness changes. Using one optimization rollout per task
($K=1$) remains 5.74 and 7.82 points below the full method, respectively,
demonstrating the benefit of repeated execution evidence.

The hard-gate-only variant evaluates nine rounds yet selects a lower-scoring
harness than full \method, which terminates in round eight. Behavioral checks
therefore complement aggregate score changes when selecting an intervention.
Appendix Figure~\ref{fig:k_repeat_count} further shows that $K=3$ gives the
strongest final harness in the measured sweep (50.9\%), while $K=5$ increases
rollout cost without improving the result (49.7\%).

\subsection{\textcolor{black}{Insights into Harness Adaptation}}

\textcolor{black}{Appendix Tables~\ref{tab:profiles} and~\ref{tab:crossmodel_modes}
show that \textbf{useful interventions are backbone dependent.} Qwen3-VL
combines code routing with action recovery and completion checks; frontier
models mainly benefit from routing, persistence, feasibility, and
completion-policy changes. OpenCUA adaptations remain close to its native
coordinate-action contract, emphasizing action normalization, file
finalization, and terminal semantics. Similar failure symptoms therefore need
not admit a universal patch.}

\textcolor{black}{Appendix Table~\ref{tab:profiles} highlights
\textbf{remaining failures that are difficult to detect from screenshots
and execution logs.}
Early edits address failures with explicit
runtime signals, such as repeated actions, invalid calls, or unsaved files.
Remaining errors increasingly involve successfully executed actions targeting
the wrong object, route, or evaluator-relevant state. These actions may appear
valid in screenshots and execution logs despite failing the task, leaving the
harness without a clear local signal to trigger recovery or guide a corrective
action. This makes further harness-only correction harder.}

\section{Related Work}
\label{sec:related}

\textbf{GUI models and runtime frameworks.}
OSWorld provides executable desktop tasks with programmatic evaluation
\citep{xie2024osworld}. Agent S/S2/S3 develop hierarchical and flat workers,
experience retrieval, specialized grounding, and coding actions
\citep{agashe2024agents,agashe2025agents2,gonzalezpumariega2025scaling};
CoAct-1 routes between GUI and programmatic execution, and OS-Symphony
coordinates planning, grounding, search, coding, and memory
\citep{song2025coact,yang2026ossymphony}. VLAA-GUI and related systems add
completion verification, loop recovery, multimodal memory, and state access
\citep{han2026vlaa,zhang2026verigui,zeng2026mementogui,yang2026stateact}.
Together they define a rich but evaluation-time-static harness design space.
We study whether execution-driven adaptation can find a better member of that
space for a given frozen model.

\textbf{GUI trajectory diagnosis and improvement.}
GUI errors unfold across state--action--state transitions. OSWorld analyzes
perception, grounding, planning, knowledge, and environment failures; CUADebug
localizes root causes from screenshots and actions; AdaMAST induces adaptive
trace taxonomies; and RoTS identifies policy errors for recovery synthesis
\citep{xie2024osworld,zhang2026cuadebug,cemri2026adamast,bu2026rots}. A
complementary reliability study uses repeated OSWorld executions to separate
environment stochasticity, instruction ambiguity, and planning variability,
establishing repeated execution as a richer account of an agent--task pair
\citep{gonzalezpumariega2026reliability}. LFF provides an early GUI-domain
attempt to turn failed executions into runtime code changes. Its paper
describes providing an LLM with the task instruction, action history, and
thought process, then deriving four families of interventions that improve
OpenCUA-72B from 42.3\% to 48.9\% \citep{sun2026learning}. \method incorporates
the screenshot sequence into the diagnostic evidence and closes the loop from
repeated task-local analysis to cross-task pattern induction, source editing,
and post-edit behavioral validation.

We defer discussions of additional related work on self-improving agents to Appendix \ref{app:additional:related:work}.

\vspace{-2mm}
\section{Conclusion}
\label{sec:conclusion}
\vspace{-2mm}

\method turns repeated multimodal execution evidence into
reusable harness changes, validated by task performance and predicted behavior.
On OSWorld-Verified, all six frozen backbones improve on held-out tasks, with
Qwen3-VL-32B-Instruct reaching 50.94\% on the full suite at 15 steps (+12.33
percentage points). It outperforms Self-Harness and Meta-Harness, and its frozen
GPT-5 harness gains 13.87 points on WindowsAgentArena at 50 steps. 
These results support harness adaptation as a mechanism for
self-improving GUI agents with frozen backbones.

\begingroup
\small
\urlstyle{same}
\bibliographystyle{iclr2027_conference}
\bibliography{references}

@inproceedings{xie2024osworld,
  title     = {{OSWorld}: Benchmarking Multimodal Agents for Open-Ended Tasks in Real Computer Environments},
  author    = {Xie, Tianbao and Zhang, Danyang and Chen, Jixuan and Li, Xiaochuan and Zhao, Siheng and Cao, Ruisheng and Hua, Toh Jing and Cheng, Zhoujun and Shin, Dongchan and Lei, Fangyu and Liu, Yitao and Xu, Yiheng and Zhou, Shuyan and Savarese, Silvio and Xiong, Caiming and Zhong, Victor and Yu, Tao},
  booktitle = {Advances in Neural Information Processing Systems},
  volume    = {37},
  pages     = {52040--52094},
  year      = {2024},
  url       = {https://arxiv.org/abs/2404.07972}
}

@article{gonzalezpumariega2025scaling,
  title   = {Scaling Agents for Computer Use},
  author  = {Gonzalez-Pumariega, Gonzalo and Tu, Vincent and Lee, Chih-Lun and Yang, Jiachen and Li, Ang and Wang, Xin Eric},
  journal = {Transactions on Machine Learning Research},
  year    = {2026},
  note    = {Originally released as arXiv:2510.02250},
  url     = {https://arxiv.org/abs/2510.02250}
}

@article{gonzalezpumariega2026reliability,
  title   = {On the Reliability of Computer Use Agents},
  author  = {Gonzalez-Pumariega, Gonzalo and Agashe, Saaket and Yang, Jiachen and Li, Ang and Wang, Xin Eric},
  journal = {arXiv preprint arXiv:2604.17849},
  year    = {2026},
  url     = {https://arxiv.org/abs/2604.17849}
}

@inproceedings{shinn2023reflexion,
  title     = {Reflexion: Language Agents with Verbal Reinforcement Learning},
  author    = {Shinn, Noah and Cassano, Federico and Gopinath, Ashwin and Narasimhan, Karthik and Yao, Shunyu},
  booktitle = {Advances in Neural Information Processing Systems},
  volume    = {36},
  pages     = {8634--8652},
  year      = {2023},
  doi       = {10.52202/075280-0377},
  url       = {https://arxiv.org/abs/2303.11366}
}

@inproceedings{zelikman2024stop,
  title     = {Self-Taught Optimizer ({STOP}): Recursively Self-Improving Code Generation},
  author    = {Zelikman, Eric and Lorch, Eliana and Mackey, Lester and Kalai, Adam Tauman},
  booktitle = {Proceedings of the First Conference on Language Modeling},
  year      = {2024},
  url       = {https://arxiv.org/abs/2310.02304}
}

@inproceedings{yang2026ossymphony,
  title     = {{OS-Symphony}: A Holistic Framework for Robust and Generalist Computer-Using Agents},
  author    = {Yang, Bowen and Jin, Kaiming and Wu, Zhenyu and Liu, Zhaoyang and Sun, Qiushi and Li, Zehao and Xie, JingJing and Liu, Zhoumianze and Xu, Fangzhi and Cheng, Kanzhi and Wang, Yian and Li, Qingyun and Qiao, Yu and Wang, Zun and Ding, Zichen},
  booktitle = {Proceedings of the 64th Annual Meeting of the Association for Computational Linguistics (Volume 1: Long Papers)},
  pages     = {22300--22330},
  year      = {2026},
  doi       = {10.18653/v1/2026.acl-long.1021},
  url       = {https://aclanthology.org/2026.acl-long.1021/}
}

@article{han2026vlaa,
  title   = {{VLAA-GUI}: Knowing When to Stop, Recover, and Search, A Modular Framework for {GUI} Automation},
  author  = {Han, Qijun and Tu, Haoqin and Wang, Zijun and Dai, Haoyue and Zhou, Yiyang and Lau, Nancy and Cardenas, Alvaro A. and Xu, Yuhui and Xu, Ran and Xiong, Caiming and Zheng, Zeyu and Yao, Huaxiu and Zhou, Yuyin and Xie, Cihang},
  journal = {arXiv preprint arXiv:2604.21375},
  year    = {2026},
  url     = {https://arxiv.org/abs/2604.21375}
}

@article{yang2026stateact,
  title   = {{StateAct}: Program State, before Pixels, for Long-Horizon Computer-Use Agents},
  author  = {Yang, Yan and Jian, Xiangru and Luo, Ziyang and Zhao, Zirui and Dai, Yutong and Shi, Ziji and Yan, Hanshu and Liew, Jun Hao and Savarese, Silvio and Li, Junnan},
  journal = {arXiv preprint arXiv:2607.22798},
  year    = {2026},
  url     = {https://arxiv.org/abs/2607.22798}
}

@article{zeng2026mementogui,
  title   = {{MementoGUI}: Learning Agentic Multimodal Memory Control for Long-Horizon {GUI} Agents},
  author  = {Zeng, Ziyun and Hua, Hang and Zou, Bocheng and Cai, Mu and Feris, Rogerio and Luo, Jiebo},
  journal = {arXiv preprint arXiv:2605.18652},
  year    = {2026},
  url     = {https://arxiv.org/abs/2605.18652}
}

@inproceedings{zhang2026verigui,
  title     = {Don't Act Blindly: Robust {GUI} Automation via Action-Effect Verification and Self-Correction},
  author    = {Zhang, Yuzhe and Xue, Xianwei and Wu, Xingyong and Chen, Mengke and Liu, Chen and He, Xinran and Shao, Run and Liu, Feiran and Xu, Huanmin and Pan, Qiutong and Wang, Haiwei},
  booktitle = {Proceedings of the 64th Annual Meeting of the Association for Computational Linguistics (Volume 1: Long Papers)},
  pages     = {28924--28941},
  year      = {2026},
  url       = {https://arxiv.org/abs/2604.05477}
}

@inproceedings{bu2026rots,
  title     = {Recovering Policy-Induced Errors: Benchmarking and Trajectory Synthesis for Robust {GUI} Agents},
  author    = {Bu, Tianpeng and Liu, Xin and Chen, Qihua and Jiang, Hao and Li, Shurui and Duan, Hongtao and Jiang, Lu and Hu, Lulu and Yang, Bin and Zhang, Minying},
  booktitle = {Proceedings of the 43rd International Conference on Machine Learning},
  year      = {2026},
  url       = {https://arxiv.org/abs/2605.29447}
}

@article{zhang2026cuadebug,
  title   = {{CUADebug}: Diagnosing and Repairing Computer-Use Agent Failures},
  author  = {Zhang, Weijia and Zhu, Kunlun and Liu, Zeyi and Chen, Yinting and Ma, Tianyi and Liu, Jiateng and Zhang, Jiaxun and Li, Bingxuan and Tang, Xiangru and Ji, Heng and You, Jiaxuan},
  journal = {arXiv preprint arXiv:2608.02643},
  year    = {2026},
  url     = {https://arxiv.org/abs/2608.02643}
}

@article{cemri2026adamast,
  title   = {Fantastic Adaptive Taxonomies and How to Use Them},
  author  = {Cemri, Mert and Cojocaru, Andrei and Pan, Melissa and Liu, Shu and Agarwal, Shubham and Krentsel, Alexander and Tang, Jay and Ramchandran, Kannan and Gonzalez, Joseph E. and Zaharia, Matei and Dimakis, Alex and Stoica, Ion},
  journal = {arXiv preprint arXiv:2607.16387},
  year    = {2026},
  url     = {https://arxiv.org/abs/2607.16387}
}

@article{sun2026learning,
  title        = {Learning from Failure: Inference-Time Self-Improvement for Computer-Use Agents},
  author       = {Sun, Xueqiao and Wang, Xiaohan and Schmidt, Ludwig and Yeung-Levy, Serena and Zhang, Yuhui},
  journal      = {arXiv preprint arXiv:2606.31270},
  year         = {2026},
  url          = {https://arxiv.org/abs/2606.31270}
}

@article{yuksekgonul2025textgrad,
  title   = {Optimizing Generative {AI} by Backpropagating Language Model Feedback},
  author  = {Yuksekgonul, Mert and Bianchi, Federico and Boen, Joseph and Liu, Sheng and Lu, Pan and Huang, Zhi and Guestrin, Carlos and Zou, James},
  journal = {Nature},
  volume  = {639},
  number  = {8055},
  pages   = {609--616},
  year    = {2025},
  publisher = {Nature Publishing Group},
  doi     = {10.1038/s41586-025-08661-4},
}

@inproceedings{pryzant2023protegi,
  title     = {Automatic Prompt Optimization with ``Gradient Descent'' and Beam Search},
  author    = {Pryzant, Reid and Iter, Dan and Li, Jerry and Lee, Yin and Zhu, Chenguang and Zeng, Michael},
  booktitle = {Proceedings of the 2023 Conference on Empirical Methods in Natural Language Processing},
  pages     = {7957--7968},
  year      = {2023},
  doi       = {10.18653/v1/2023.emnlp-main.494},
  url       = {https://aclanthology.org/2023.emnlp-main.494/}
}

@inproceedings{agrawal2025gepa,
  title     = {{GEPA}: Reflective Prompt Evolution Can Outperform Reinforcement Learning},
  author    = {Agrawal, Lakshya A. and Tan, Shangyin and Soylu, Dilara and Ziems, Noah and Khare, Rishi and Opsahl-Ong, Krista and Singhvi, Arnav and Shandilya, Herumb and Ryan, Michael J. and Jiang, Meng and others},
  booktitle = {International Conference on Learning Representations},
  year      = {2026},
  url       = {https://arxiv.org/abs/2507.19457}
}

@article{lee2025feedbackdescent,
  title   = {Feedback Descent: Open-Ended Text Optimization via Pairwise Comparison},
  author  = {Lee, Yoonho and Boen, Joseph and Finn, Chelsea},
  journal = {arXiv preprint arXiv:2511.07919},
  year    = {2025},
  url     = {https://arxiv.org/abs/2511.07919}
}

@article{lee2026metaharness,
  title   = {Meta-Harness: End-to-End Optimization of Model Harnesses},
  author  = {Lee, Yoonho and Nair, Roshen and Zhang, Qizheng and Lee, Kangwook and Khattab, Omar and Finn, Chelsea},
  journal = {arXiv preprint arXiv:2603.28052},
  year    = {2026},
  url     = {https://arxiv.org/abs/2603.28052}
}

@article{lin2026ahe,
  title   = {Agentic Harness Engineering: Observability-Driven Automatic Evolution of Coding-Agent Harnesses},
  author  = {Lin, Jiahang and Liu, Shichun and Pan, Chengjun and Lin, Lizhi and Dou, Shihan and Xi, Zhiheng and Huang, Xuanjing and Yan, Hang and Han, Zhenhua and Gui, Tao and Jiang, Yu-Gang},
  journal = {arXiv preprint arXiv:2604.25850},
  year    = {2026},
  url     = {https://arxiv.org/abs/2604.25850}
}

@article{zhang2026selfharness,
  title   = {Self-Harness: Harnesses That Improve Themselves},
  author  = {Zhang, Hangfan and Zhang, Shao and Li, Kangcong and Zhang, Chen and Chen, Yang and Zhang, Yiqun and Bai, Lei and Hu, Shuyue},
  journal = {arXiv preprint arXiv:2606.09498},
  year    = {2026},
  url     = {https://arxiv.org/abs/2606.09498}
}

@article{chen2026harnessfix,
  title   = {From Failed Trajectories to Reliable {LLM} Agents: Diagnosing and Repairing Harness Flaws},
  author  = {Chen, Mengzhuo and Wang, Junjie and Liu, Zhe and Wang, Yawen and Zheng, Haiming and Wang, Qing},
  journal = {arXiv preprint arXiv:2606.06324},
  year    = {2026},
  url     = {https://arxiv.org/abs/2606.06324}
}

@article{park2026autosaddler,
  title   = {{AutoSaddler}: Automatic Harness Optimization with Durable Updates from Agent Execution Traces},
  author  = {Park, Sungho and Kim, Wonjoong and Tan, Rongyuan and Zhang, Jue and Han, Wook-Shin and Gao, Pengfei and Park, Chanyoung and Yao, Yongqiang and Fu, Rao and Nallipogu, Elsie and Lin, Qingwei and Rajmohan, Saravan and Zhang, Dongmei},
  journal = {arXiv preprint arXiv:2608.23041},
  year    = {2026},
  url     = {https://arxiv.org/abs/2608.23041}
}

@article{luo2026harnessbank,
  title   = {{HarnessBank}: Semantic Gene-Bank Search with Gated Verification for Agent-Harness Self-Evolution},
  author  = {Luo, Xiaotian and Xue, Dizhan and Wang, Fengxingyu and Hu, Chuanrui and Deng, Yafeng},
  journal = {arXiv preprint arXiv:2607.13683},
  year    = {2026},
  url     = {https://arxiv.org/abs/2607.13683}
}

@article{zhang2026darwinx,
  title   = {{DarwinX}: Evolving Agent Harnesses Through Natural Selection},
  author  = {Zhang, Yifan and Dai, Yutong and Tan, Juntao and Yang, Luyu and Mullur, Rishi and Hoang, Thai and Hu, Zhiyuan and Zhu, James and Mui, Phil and Savarese, Silvio and Xu, Ran and Chen, Zeyuan},
  journal = {arXiv preprint arXiv:2608.07545},
  year    = {2026},
  url     = {https://arxiv.org/abs/2608.07545}
}

@article{shao2026harnessr1,
  title   = {{Harness-R1}: Learning to Edit Executable Runtime Harnesses from Agent Failure Trajectories},
  author  = {Shao, Shuai and Zhang, Kangning and Li, Qingyao and Wang, Shijian and Wang, Hao and Jiao, Wenxiang and Lu, Yuan and Guo, Yi and Liu, Weiwen and Zhang, Weinan},
  journal = {arXiv preprint arXiv:2608.02276},
  year    = {2026},
  url     = {https://arxiv.org/abs/2608.02276}
}

@inproceedings{agashe2024agents,
  title     = {{Agent S}: An Open Agentic Framework that Uses Computers Like a Human},
  author    = {Agashe, Saaket and Han, Jiuzhou and Gan, Shuyu and Yang, Jiachen and Li, Ang and Wang, Xin Eric},
  booktitle = {International Conference on Learning Representations},
  year      = {2025},
  url       = {https://arxiv.org/abs/2410.08164}
}

@inproceedings{agashe2025agents2,
  title     = {{Agent S2}: A Compositional Generalist-Specialist Framework for Computer Use Agents},
  author    = {Agashe, Saaket and Wong, Kyle and Tu, Vincent and Yang, Jiachen and Li, Ang and Wang, Xin Eric},
  booktitle = {Conference on Language Modeling},
  year      = {2025},
  url       = {https://openreview.net/forum?id=zg5is4GJ3R}
}

@inproceedings{song2025coact,
  title     = {{CoAct-1}: Computer-using Multi-Agent System with Coding Actions},
  author    = {Song, Linxin and Dai, Yutong and Prabhu, Viraj and Zhang, Jieyu and Shi, Taiwei and Li, Li and Li, Junnan and Savarese, Silvio and Chen, Zeyuan and Zhao, Jieyu and Xu, Ran and Xiong, Caiming},
  booktitle = {International Conference on Learning Representations},
  year      = {2026},
  url       = {https://openreview.net/forum?id=l1MQVgIKEU}
}

@inproceedings{yang2025gta1,
  title     = {{GTA1}: {GUI} Test-time Scaling Agent},
  author    = {Yang, Yan and Li, Dongxu and Dai, Yutong and Yang, Yuhao and Luo, Ziyang and Zhao, Zirui and Hu, Zhiyuan and Huang, Junzhe and Saha, Amrita and Chen, Zeyuan and others},
  booktitle = {International Conference on Learning Representations},
  year      = {2026},
  url       = {https://arxiv.org/abs/2507.05791}
}

@article{qin2025uitars,
  title   = {{UI-TARS}: Pioneering Automated {GUI} Interaction with Native Agents},
  author  = {Qin, Yujia and Ye, Yining and Fang, Junjie and Wang, Haoming and Liang, Shihao and Tian, Shizuo and Zhang, Junda and Li, Jiahao and Li, Yunxin and Huang, Shijue and others},
  journal = {arXiv preprint arXiv:2501.12326},
  year    = {2025},
  url     = {https://arxiv.org/abs/2501.12326}
}

@inproceedings{wang2025opencua,
  title     = {{OpenCUA}: Open Foundations for Computer-Use Agents},
  author    = {Wang, Xinyuan and Wang, Bowen and Lu, Dunjie and Yang, Junlin and Xie, Tianbao and Wang, Junli and Deng, Jiaqi and Guo, Xiaole and Xu, Yiheng and Wu, Chen and others},
  booktitle = {Advances in Neural Information Processing Systems},
  volume    = {38},
  pages     = {139756--139806},
  year      = {2025},
  doi       = {10.52202/085713-4669},
  url       = {https://proceedings.neurips.cc/paper_files/paper/2025/hash/cc7ae529e945226b0d52ea4ac478c4f3-Abstract-Conference.html}
}

@article{qwen2025qwen3vl,
  title   = {{Qwen3-VL} Technical Report},
  author  = {Bai, Shuai and Cai, Yuxuan and Chen, Ruizhe and Chen, Keqin and Chen, Xionghui and others},
  journal = {arXiv preprint arXiv:2511.21631},
  year    = {2025},
  url     = {https://arxiv.org/abs/2511.21631}
}

@article{wang2026harnesslens,
  title   = {Verify Smarter, Evolve Further: Efficient Harness Evolution through Behavior-Aware Verification},
  author  = {Xu, Jinghan and Zhang, Yikai and Chen, Aili and Li, Weiyuan and Liang, Jiaqing and Yang, Deqing},
  journal = {arXiv preprint arXiv:2608.27311},
  year    = {2026},
  url     = {https://arxiv.org/abs/2608.27311}
}

@article{liu2026harnessevolve,
  title   = {{HarnessEvolve}: Learning from Reference Trajectories for Reliable Agent Self-Evolution},
  author  = {Jiang, Wen and Chu, Mingmin and Tian, Yimeng and Zhang, Qianxin and Yang, Haofei and Yang, Rui and Liu, Yang and Lv, Tao and Li, Fangming},
  journal = {arXiv preprint arXiv:2609.00829},
  year    = {2026},
  url     = {https://arxiv.org/abs/2609.00829}
}

@inproceedings{bonatti2025windows,
  title     = {Windows Agent Arena: Evaluating Multi-Modal {OS} Agents at Scale},
  author    = {Bonatti, Rogerio and Zhao, Dan and Bonacci, Francesco and Dupont, Dillon and Abdali, Sara and Li, Yinheng and Lu, Yadong and Wagle, Justin and Koishida, Kazuhito and Bucker, Arthur and Jang, Lawrence Keunho and Hui, Zheng},
  booktitle = {Proceedings of the 42nd International Conference on Machine Learning},
  volume    = {267},
  series    = {Proceedings of Machine Learning Research},
  pages     = {4874--4910},
  year      = {2025},
  publisher = {PMLR},
  url       = {https://proceedings.mlr.press/v267/bonatti25a.html}
}

@misc{osworldverifiedleaderboard2026,
  title        = {{OSWorld-Verified Results}},
  author       = {{OSWorld Team}},
  year         = {2026},
  howpublished = {Official benchmark leaderboard},
  url          = {https://osworld-v1.xlang.ai/},
  note         = {Accessed 2026-09-23}
}

@misc{wei2026steplevel,
  title         = {Step-level Optimization for Efficient Computer-use Agents},
  author        = {Wei, Jinbiao and Ni, Kangqi and Zhao, Yilun and Gan, Guo and Cohan, Arman},
  year          = {2026},
  eprint        = {2604.27151},
  archivePrefix = {arXiv},
  primaryClass  = {cs.AI},
  url           = {https://arxiv.org/abs/2604.27151}
}
\endgroup

\clearpage
\appendix
\section{Additional Related Work}
\label{app:additional:related:work}
\paragraph{Self-improving agents and automatic harness optimization.}
Self-improving agent systems retain execution feedback in forms that influence
later behavior. Reflexion stores linguistic feedback in episodic memory, while
STOP improves an executable scaffolding program
\citep{shinn2023reflexion,zelikman2024stop}.
TextGrad, ProTeGi, GEPA, and Feedback Descent optimize textual components from
language feedback \citep{yuksekgonul2025textgrad,pryzant2023protegi,
agrawal2025gepa,lee2025feedbackdescent}. Meta-Harness and Self-Harness expand
the editable object to agent source code and retain changes using execution
results. Meta-Harness evaluates classification, mathematical reasoning, and
terminal coding, while Self-Harness evaluates Terminal-Bench, SWE-bench, and
AppWorld \citep{lee2026metaharness,zhang2026selfharness}. Their
optimizer-facing evidence consists predominantly of serialized source,
language/tool traces, and scores. Concurrent systems
study repair, durable update archives, learned editors, and behavior-aware
validation \citep{lin2026ahe,chen2026harnessfix,park2026autosaddler,
luo2026harnessbank,zhang2026darwinx,shao2026harnessr1,
wang2026harnesslens,liu2026harnessevolve}. \method specializes this search to
GUI interaction, evolving the executable harness from same-task repeated visual
evidence and selecting updates through held-out utility and predicted-behavior
validation.

\section{Extended OSWorld Comparisons}
\label{app:additional}

\subsection{Complete split-level adaptation results}

Table~\ref{tab:cross_backbone_full} reports the Search and Validation results
used during optimization alongside the Test and Full operating
points summarized in the main paper.

\begin{table}[H]
\centering
\caption{Complete paired 15-step results before and after harness optimization.
Search and Validation guide optimization, whereas Test remains sealed until the
final harness is selected. Full reports all 361 OSWorld-Verified tasks.
Parentheses show absolute percentage-point gains over the corresponding
initial harness.}
\label{tab:cross_backbone_full}
\scriptsize
\setlength{\tabcolsep}{2.0pt}
\renewcommand{\arraystretch}{1.18}
\definecolor{harvestTestBg}{HTML}{EEF3F8}
\definecolor{harvestFullBg}{HTML}{F7F1E5}
\begin{tabular*}{\textwidth}{@{\extracolsep{\fill}}lrr
>{\columncolor{harvestTestBg}}r>{\columncolor{harvestFullBg}}r
rr>{\columncolor{harvestTestBg}}r
>{\columncolor{harvestFullBg}[\tabcolsep][0pt]}r@{}}
\toprule
& \multicolumn{4}{c}{Initial harness $H_0$} & \multicolumn{4}{c}{\method $H^*$} \\
\cmidrule(lr){2-5}\cmidrule(lr){6-9}
Backbone & Search & Val. & Test & Full & Search & Val. & Test & Full \\
\midrule
Qwen3-VL-8B-Instruct & 31.01 & 29.72 & 28.79 & 29.49 & 41.43 {\color{green!50!black}(+10.41)} & 37.22 {\color{green!50!black}(+7.50)} & \textbf{34.84} {\color{green!50!black}(+6.05)} & \textbf{36.83} {\color{green!50!black}(+7.34)} \\
Qwen3-VL-32B-Instruct & 31.07 & 35.06 & 43.02 & 38.61 & 46.25 {\color{green!50!black}(+15.18)} & 50.00 {\color{green!50!black}(+14.94)} & \textbf{53.18} {\color{green!50!black}(+10.16)} & \textbf{50.94} {\color{green!50!black}(+12.33)} \\
OpenCUA-32B & 27.40 & 28.75 & 32.82 & 30.71 & 33.69 {\color{green!50!black}(+6.29)} & 34.63 {\color{green!50!black}(+5.88)} & \textbf{34.31} {\color{green!50!black}(+1.49)} & \textbf{34.24} {\color{green!50!black}(+3.53)} \\
OpenCUA-72B & 36.25 & 30.00 & 41.25 & 37.65 & 42.50 {\color{green!50!black}(+6.25)} & 37.50 {\color{green!50!black}(+7.50)} & \textbf{44.18} {\color{green!50!black}(+2.93)} & \textbf{42.33} {\color{green!50!black}(+4.68)} \\
Gemini 3.1 Pro & 65.33 & 67.69 & 66.42 & 66.46 & 72.68 {\color{green!50!black}(+7.35)} & 75.92 {\color{green!50!black}(+8.23)} & \textbf{72.63} {\color{green!50!black}(+6.20)} & \textbf{73.37} {\color{green!50!black}(+6.91)} \\
GPT-5 & 52.50 & 55.00 & 56.35 & 55.20 & 66.25 {\color{green!50!black}(+13.75)} & 60.00 {\color{green!50!black}(+5.00)} & \textbf{61.86} {\color{green!50!black}(+5.51)} & \textbf{62.42} {\color{green!50!black}(+7.22)} \\
\bottomrule
\end{tabular*}
\end{table}

\subsection{Benchmark visualization and domain-level results}

Tables~\ref{tab:benchmark_qwen}--
\ref{tab:benchmark_frontier} give the corresponding domain-level results and a
broader accounting of public runs involving our target backbones. Published
rows retain their original model, runtime, step budget, and aggregation
protocol; domain entries remain blank when the source reports only an overall
score.

\begingroup
\scriptsize
\setlength{\tabcolsep}{2.35pt}
\renewcommand{\arraystretch}{1.01}
\setlength{\LTcapwidth}{\textwidth}
\setlength{\LTleft}{0pt}
\setlength{\LTright}{0pt}
\begin{longtable}{@{\extracolsep{\fill}}p{0.43\textwidth}r*{6}{r}@{}}
\caption{OSWorld-Verified comparison with general open-weight backbones. Public
rows retain their original runtime and evaluation budget; GUI-HARVEST is
optimized with 15-step Search rollouts and frozen before evaluation at all
three budgets~\citep{xie2024osworld,yang2026ossymphony,qwen2025qwen3vl,osworldverifiedleaderboard2026}.
``--'' denotes an unreported domain score; bold marks the best reported value
within each step-budget block.}\label{tab:benchmark_qwen}\label{tab:benchmark_full}\\
\toprule
Method / model & Step & OS & Office & Daily & Prof. & Multi & Avg. \\
\midrule
\endfirsthead
\multicolumn{8}{l}{\scriptsize\itshape Table~\ref{tab:benchmark_qwen} continued.}\\
\toprule
Method / model & Step & OS & Office & Daily & Prof. & Multi & Avg. \\
\midrule
\endhead
\midrule
\multicolumn{8}{r}{\scriptsize\itshape Continued on next page.}\\
\endfoot
\bottomrule
\endlastfoot

\rowcolor{gray!16}\multicolumn{8}{c}{\itshape Max 15 Steps}\\
GUI-HARVEST / Qwen3-VL-8B-Instruct & 15 & 58.33 & 32.47 & 38.64 & 67.35 & 19.16 & 36.83 \\
GUI-HARVEST / Qwen3-VL-32B-Instruct & 15 & \textbf{83.33} & \textbf{48.68} & \textbf{57.62} & \textbf{75.51} & \textbf{26.88} & \textbf{50.94} \\

\midrule
\rowcolor{gray!16}\multicolumn{8}{c}{\itshape Max 50 Steps}\\
Qwen / Qwen3-VL-8B-Instruct~\citep{yang2026ossymphony} & 50 & -- & -- & -- & -- & -- & 33.90 \\
OS-Symphony / Qwen3-VL-8B-Instruct~\citep{yang2026ossymphony} & 50 & -- & -- & -- & -- & -- & 33.90 \\
Qwen / Qwen3-VL-32B-Instruct~\citep{yang2026ossymphony} & 50 & -- & -- & -- & -- & -- & 32.60 \\
Agent S3 / Qwen3-VL-32B-Instruct~\citep{yang2026ossymphony} & 50 & 50.00 & 36.67 & 50.62 & 61.22 & 21.96 & 40.11 \\
Qwen / Qwen3-VL-32B-Thinking~\citep{yang2026ossymphony} & 50 & -- & -- & -- & -- & -- & 41.00 \\
OS-Symphony / Qwen3-VL-32B-Instruct~\citep{yang2026ossymphony} & 50 & 58.33 & 40.94 & 53.54 & 75.10 & 31.24 & 46.86 \\
\midrule
GUI-HARVEST / Qwen3-VL-8B-Instruct & 50 & 54.17 & 32.47 & 45.61 & 59.18 & 24.26 & 38.26 \\
GUI-HARVEST / Qwen3-VL-32B-Instruct & 50 & \textbf{75.00} & \textbf{46.18} & \textbf{60.08} & \textbf{75.51} & \textbf{32.35} & \textbf{51.52} \\

\midrule
\rowcolor{gray!16}\multicolumn{8}{c}{\itshape Max 100 Steps}\\
Qwen / Qwen2.5-VL-32B-Instruct~\citep{osworldverifiedleaderboard2026} & 100 & -- & -- & -- & -- & -- & 3.88 \\
Qwen / Qwen2.5-VL-72B-Instruct~\citep{osworldverifiedleaderboard2026} & 100 & -- & -- & -- & -- & -- & 5.00 \\
\midrule
GUI-HARVEST / Qwen3-VL-8B-Instruct & 100 & 50.00 & 33.33 & 48.17 & 55.10 & 28.57 & 39.37 \\
GUI-HARVEST / Qwen3-VL-32B-Instruct & 100 & \textbf{70.83} & \textbf{46.18} & \textbf{61.37} & \textbf{75.51} & \textbf{33.42} & \textbf{51.79} \\
\end{longtable}
\endgroup

\begingroup
\scriptsize
\setlength{\tabcolsep}{2.35pt}
\renewcommand{\arraystretch}{1.01}
\setlength{\LTcapwidth}{\textwidth}
\setlength{\LTleft}{0pt}
\setlength{\LTright}{0pt}
\begin{longtable}{@{\extracolsep{\fill}}p{0.43\textwidth}r*{6}{r}@{}}
\caption{OSWorld-Verified comparison with GUI-specialized open models.
OpenCUA rows use the official coordinate-action runtime; LFF and GUI-HARVEST
modify that runtime family~\citep{qin2025uitars,wang2025opencua,sun2026learning,yang2026ossymphony,osworldverifiedleaderboard2026}.
``--'' denotes an unreported domain score; bold marks the best reported value
within each step-budget block.}\label{tab:benchmark_opencua}\\
\toprule
Method / model & Step & OS & Office & Daily & Prof. & Multi & Avg. \\
\midrule
\endfirsthead
\multicolumn{8}{l}{\scriptsize\itshape Table~\ref{tab:benchmark_opencua} continued.}\\
\toprule
Method / model & Step & OS & Office & Daily & Prof. & Multi & Avg. \\
\midrule
\endhead
\midrule
\multicolumn{8}{r}{\scriptsize\itshape Continued on next page.}\\
\endfoot
\bottomrule
\endlastfoot

\rowcolor{gray!16}\multicolumn{8}{c}{\itshape Max 15 Steps}\\
UI-TARS-1.5-7B~\citep{song2025coact} & 15 & 34.78 & 27.19 & 27.99 & 61.45 & 5.38 & 25.76 \\
OpenCUA / OpenCUA-32B~\citep{wang2025opencua} & 15 & -- & -- & -- & -- & -- & 29.71 \\
OpenCUA / OpenCUA-72B~\citep{wang2025opencua} & 15 & -- & -- & -- & -- & -- & 39.03 \\
\midrule
GUI-HARVEST / OpenCUA-32B & 15 & \textbf{58.33} & 29.05 & 44.76 & 65.31 & 9.36 & 34.24 \\
GUI-HARVEST / OpenCUA-72B & 15 & 45.83 & \textbf{41.01} & \textbf{48.67} & \textbf{75.51} & \textbf{20.28} & \textbf{42.33} \\

\midrule
\rowcolor{gray!16}\multicolumn{8}{c}{\itshape Max 50 Steps}\\
UI-TARS-1.5-7B~\citep{song2025coact} & 50 & 25.00 & 26.51 & 31.41 & 48.91 & 9.77 & 25.08 \\
OpenCUA / OpenCUA-32B~\citep{wang2025opencua} & 50 & -- & -- & -- & -- & -- & 34.20 \\
OpenCUA / OpenCUA-72B~\citep{wang2025opencua} & 50 & -- & -- & -- & -- & -- & 44.89 \\
\midrule
GUI-HARVEST / OpenCUA-32B & 50 & 66.67 & 31.62 & \textbf{51.18} & 65.31 & 13.66 & 38.12 \\
GUI-HARVEST / OpenCUA-72B & 50 & \textbf{70.83} & \textbf{47.85} & 46.10 & \textbf{71.43} & \textbf{26.72} & \textbf{46.76} \\

\midrule
\rowcolor{gray!16}\multicolumn{8}{c}{\itshape Max 100 Steps}\\
UI-TARS-1.5-7B~\citep{song2025coact} & 100 & 29.17 & 25.01 & 31.07 & 46.99 & 8.80 & 25.41 \\
OpenCUA / OpenCUA-32B~\citep{wang2025opencua} & 100 & -- & -- & -- & -- & -- & 34.88 \\
LFF / OpenCUA-32B~\citep{sun2026learning} & 100 & -- & -- & -- & -- & -- & 38.20 \\
DeepMiner-Mano-7B~\citep{yang2026ossymphony} & 100 & 50.00 & 39.28 & 44.87 & 73.47 & 17.20 & 40.15 \\
UI-TARS~\citep{yang2026ossymphony} & 100 & 41.67 & 50.42 & 55.69 & 51.02 & 14.66 & 41.85 \\
OpenCUA / OpenCUA-72B (three-run mean)~\citep{wang2025opencua} & 100 & -- & -- & -- & -- & -- & 44.99 \\
OpenCUA / OpenCUA-72B (leaderboard run)~\citep{han2026vlaa} & 100 & 61.13 & 44.73 & 49.95 & 72.58 & 22.16 & 44.91 \\
LFF / OpenCUA-72B~\citep{sun2026learning} & 100 & -- & -- & -- & -- & -- & $48.9\!\pm\!1.2$ \\
UI-TARS-2~\citep{yang2026ossymphony} & 100 & 41.67 & 61.11 & \textbf{62.12} & 61.22 & 34.13 & 53.10 \\
DeepMiner-Mano-72B~\citep{yang2026ossymphony} & 100 & 66.67 & \textbf{63.22} & 52.51 & \textbf{83.67} & 24.41 & \textbf{53.91} \\
\midrule
GUI-HARVEST / OpenCUA-32B & 100 & \textbf{70.83} & 33.33 & 51.18 & 65.31 & 15.81 & 39.50 \\
GUI-HARVEST / OpenCUA-72B & 100 & \textbf{70.83} & 50.41 & 44.82 & 71.43 & \textbf{34.25} & 49.25 \\
\end{longtable}
\endgroup

\begingroup
\scriptsize
\setlength{\tabcolsep}{2.35pt}
\renewcommand{\arraystretch}{1.01}
\setlength{\LTcapwidth}{\textwidth}
\setlength{\LTleft}{0pt}
\setlength{\LTright}{0pt}
\begin{longtable}{@{\extracolsep{\fill}}p{0.43\textwidth}r*{6}{r}@{}}
\caption{OSWorld-Verified comparison with proprietary frontier backbones and
systems. Each public row preserves its original framework and step budget and
provides literature-level benchmark context~\citep{agashe2025agents2,song2025coact,yang2025gta1,gonzalezpumariega2025scaling,yang2026ossymphony,han2026vlaa,osworldverifiedleaderboard2026}.
``--'' denotes an unreported domain score; bold marks the best reported value
within each step-budget block.}\label{tab:benchmark_frontier}\\
\toprule
Method / model & Step & OS & Office & Daily & Prof. & Multi & Avg. \\
\midrule
\endfirsthead
\multicolumn{8}{l}{\scriptsize\itshape Table~\ref{tab:benchmark_frontier} continued.}\\
\toprule
Method / model & Step & OS & Office & Daily & Prof. & Multi & Avg. \\
\midrule
\endhead
\midrule
\multicolumn{8}{r}{\scriptsize\itshape Continued on next page.}\\
\endfoot
\bottomrule
\endlastfoot

\rowcolor{gray!16}\multicolumn{8}{c}{\itshape Max 15 Steps}\\
OpenAI o3~\citep{song2025coact} & 15 & 37.50 & 1.45 & 8.02 & 12.29 & 11.82 & 9.09 \\
OpenAI CUA / GPT-4o~\citep{song2025coact} & 15 & 45.83 & 22.17 & 37.65 & 41.22 & 10.75 & 26.01 \\
Jedi-7B w/ GPT-4o~\citep{osworldverifiedleaderboard2026} & 15 & -- & -- & -- & -- & -- & 26.80 \\
Agent S2.5 / OpenAI o3~\citep{song2025coact} & 15 & 70.83 & 42.85 & 44.61 & 57.10 & 17.82 & 38.98 \\
CoAct-1 / GPT-5~\citep{song2025coact} & 15 & 66.67 & 47.18 & 42.30 & 47.74 & 23.82 & 39.81 \\
VLAA-GUI / Gemini 3 Flash~\citep{han2026vlaa} & 15 & 79.20 & 29.92 & 54.14 & 57.13 & 34.00 & 43.15 \\
VLAA-GUI / Gemini 3.1 Pro~\citep{han2026vlaa} & 15 & 83.30 & 52.99 & 54.33 & 61.22 & 34.70 & 51.69 \\
VLAA-GUI / Claude Sonnet 4.6~\citep{han2026vlaa} & 15 & 83.30 & 69.72 & 58.72 & 57.13 & \textbf{60.20} & 64.13 \\
VLAA-GUI / Claude Opus 4.6~\citep{han2026vlaa} & 15 & 83.30 & 60.65 & 66.38 & 79.59 & 55.90 & 64.75 \\
\midrule
GUI-HARVEST / Gemini 3.1 Pro & 15 & \textbf{87.50} & \textbf{81.73} & \textbf{67.71} & \textbf{87.76} & 56.36 & \textbf{73.37} \\
GUI-HARVEST / GPT-5 & 15 & 83.33 & 65.78 & 61.09 & 73.47 & 48.10 & 62.42 \\

\midrule
\rowcolor{gray!16}\multicolumn{8}{c}{\itshape Max 50 Steps}\\
OpenAI o3~\citep{song2025coact} & 50 & 37.50 & 11.50 & 19.78 & 30.10 & 11.82 & 17.17 \\
OpenAI CUA / GPT-4o~\citep{song2025coact} & 50 & 70.83 & 23.56 & 38.43 & 52.09 & 15.86 & 31.19 \\
Claude 3.7 Sonnet~\citep{osworldverifiedleaderboard2026} & 50 & -- & -- & -- & -- & -- & 35.80 \\
Agent S3 / GPT-5-Mini~\citep{yang2026ossymphony} & 50 & 62.50 & 54.62 & 46.67 & 44.90 & 37.04 & 47.58 \\
UiPath Screen Agent / GPT-5~\citep{yang2026ossymphony} & 50 & 73.91 & 49.52 & 62.12 & 71.43 & 37.30 & 53.69 \\
Agent S2.5 / OpenAI o3~\citep{song2025coact} & 50 & 75.00 & 52.81 & 55.80 & 75.42 & 39.53 & 54.21 \\
CoAct-1 / GPT-5~\citep{yang2026ossymphony} & 50 & 70.83 & 60.65 & 54.09 & 69.39 & 42.37 & 56.39 \\
OS-Symphony / GPT-5-Mini~\citep{yang2026ossymphony} & 50 & 73.68 & 58.17 & 61.39 & 75.00 & 47.37 & 58.05 \\
Claude Sonnet 4.5~\citep{osworldverifiedleaderboard2026} & 50 & -- & -- & -- & -- & -- & 58.08 \\
Agent S3 / GPT-5~\citep{gonzalezpumariega2025scaling} & 50 & -- & -- & -- & -- & -- & 61.10 \\
VLAA-GUI / Gemini 3 Flash~\citep{han2026vlaa} & 50 & 83.30 & 69.19 & 62.75 & 62.49 & 51.00 & 63.14 \\
OS-Symphony / GPT-5~\citep{yang2026ossymphony} & 50 & 75.00 & 64.85 & 61.19 & 69.23 & 54.86 & 63.61 \\
UiPath Screen Agent / Claude Opus 4.5~\citep{osworldverifiedleaderboard2026} & 50 & -- & -- & -- & -- & -- & 64.40 \\
VLAA-GUI / Gemini 3.1 Pro~\citep{han2026vlaa} & 50 & 83.30 & 71.51 & 67.32 & 67.35 & 55.90 & 66.80 \\
VLAA-GUI / Claude Sonnet 4.6~\citep{han2026vlaa} & 50 & 83.30 & 79.23 & 69.24 & 57.13 & \textbf{66.70} & 71.11 \\
VLAA-GUI / Claude Opus 4.6~\citep{han2026vlaa} & 50 & 83.30 & 78.60 & 72.60 & 83.67 & 61.30 & 73.85 \\
\midrule
GUI-HARVEST / Gemini 3.1 Pro & 50 & \textbf{91.67} & \textbf{86.00} & \textbf{72.84} & \textbf{87.76} & 63.73 & \textbf{78.03} \\
GUI-HARVEST / GPT-5 & 50 & 87.50 & 69.77 & 64.93 & 73.47 & 52.40 & 65.93 \\

\midrule
\rowcolor{gray!16}\multicolumn{8}{c}{\itshape Max 100 Steps}\\
OpenAI o3~\citep{song2025coact} & 100 & 62.50 & 17.23 & 26.29 & 38.79 & 16.53 & 23.00 \\
Jedi-7B w/ GPT-4o~\citep{osworldverifiedleaderboard2026} & 100 & -- & -- & -- & -- & -- & 29.30 \\
Qwen3-VL API / Qwen3-VL-Flash~\citep{osworldverifiedleaderboard2026} & 100 & -- & -- & -- & -- & -- & 41.57 \\
Agent S2.5 / GPT-5~\citep{yang2025gta1} & 100 & -- & -- & -- & -- & -- & 58.40 \\
CoAct-1 / GPT-5~\citep{han2026vlaa} & 100 & 75.00 & 62.93 & 57.94 & 71.43 & 47.87 & 59.93 \\
Seed / Seed-1.8~\citep{han2026vlaa} & 100 & 66.67 & 68.80 & 67.05 & 71.43 & 42.38 & 61.87 \\
Agent S3 / GPT-5~\citep{yang2026ossymphony} & 100 & 77.50 & 66.46 & 61.23 & 69.80 & 51.37 & 62.63 \\
Claude Sonnet 4.5~\citep{han2026vlaa} & 100 & 70.83 & 72.59 & 61.35 & 63.27 & 49.54 & 62.84 \\
GTA1 / GPT-5~\citep{han2026vlaa} & 100 & 79.17 & 63.91 & 62.56 & 79.59 & 50.91 & 63.41 \\
OS-Symphony / GPT-5~\citep{yang2026ossymphony} & 100 & 79.17 & 65.73 & 67.76 & 69.23 & 57.98 & 65.84 \\
UiPath Screen Agent / Claude Opus 4.5~\citep{han2026vlaa} & 100 & 70.83 & 74.13 & 68.33 & 73.47 & 52.97 & 67.14 \\
Agent S3 / Claude Opus 4.5~\citep{han2026vlaa} & 100 & 75.00 & 76.06 & 67.51 & 59.18 & 59.00 & 67.46 \\
VLAA-GUI / Gemini 3 Flash~\citep{han2026vlaa} & 100 & \textbf{91.70} & 64.90 & 74.85 & 67.33 & 63.40 & 68.77 \\
VLAA-GUI / Claude Sonnet 4.6~\citep{han2026vlaa} & 100 & 83.30 & 79.24 & 69.25 & 57.14 & \textbf{68.80} & 71.67 \\
VLAA-GUI / Gemini 3.1 Pro~\citep{han2026vlaa} & 100 & 83.30 & 76.60 & 73.76 & 73.47 & 62.90 & 72.47 \\
HIPPO / Claude Opus 4.5~\citep{han2026vlaa} & 100 & 87.50 & 74.27 & 69.27 & \textbf{95.92} & 64.31 & 74.49 \\
VLAA-GUI / Claude Opus 4.6~\citep{han2026vlaa} & 100 & \textbf{91.70} & 82.87 & \textbf{75.17} & 83.67 & 65.60 & 77.45 \\
OpenAPA / Gemini 3.1 Pro~\citep{osworldverifiedleaderboard2026} & 100 & -- & -- & -- & -- & -- & 78.34 \\
\midrule
GUI-HARVEST / Gemini 3.1 Pro & 100 & 91.67 & \textbf{86.86} & 74.12 & 87.76 & 65.88 & \textbf{79.14} \\
GUI-HARVEST / GPT-5 & 100 & 87.50 & 73.18 & 66.22 & 75.51 & 55.63 & 68.42 \\
\end{longtable}
\endgroup

\begin{figure}[H]
  \centering
  \includegraphics[width=\textwidth]{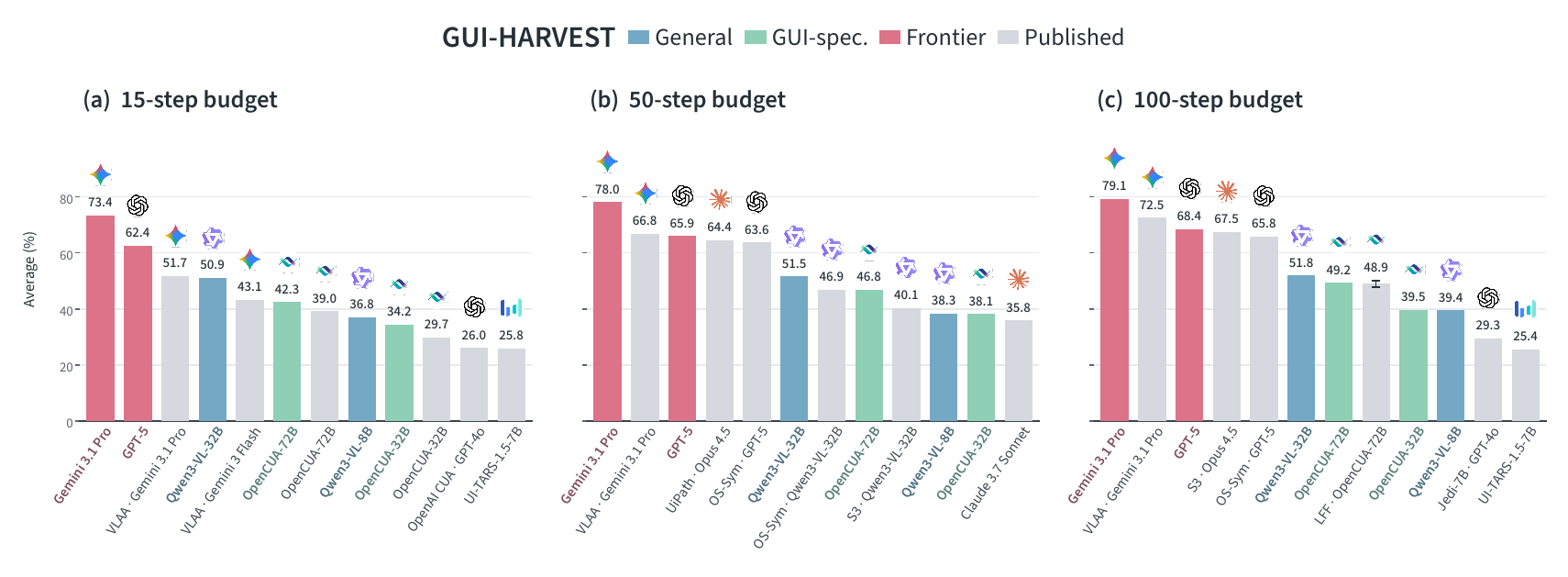}
  \caption{Selected OSWorld-Verified operating points at 15, 50, and 100
  maximum environment steps. Colored bars show frozen GUI-HARVEST harnesses
  optimized with 15-step Search rollouts; gray bars show published systems in
  their reported model and runtime configurations. This visualization
  summarizes the domain-level tables above; published values are drawn from
  the corresponding papers and the official OSWorld leaderboard
  \citep{song2025coact,wang2025opencua,sun2026learning,yang2026ossymphony,han2026vlaa,osworldverifiedleaderboard2026}.}
  \label{fig:benchmark_bars}
\end{figure}

\section{Repeated Execution and Repeat-Count Ablation}
\label{app:repeated}

\begin{figure}[H]
  \centering
  \includegraphics[width=\textwidth]{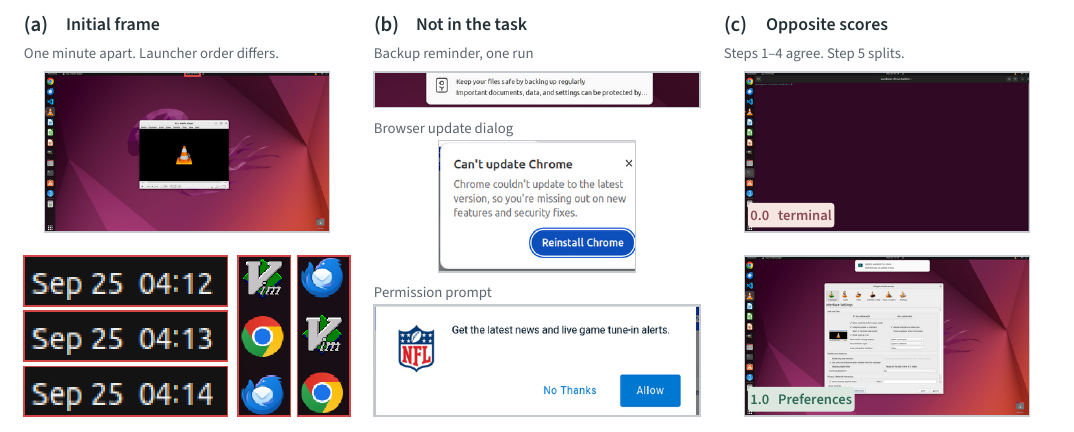}
  \caption{Execution variability in desktop GUI tasks. (a) Clean resets one
  minute apart already differ in the initial frame. (b) Task-irrelevant system
  and web events enter the observation stream during execution. (c) Two runs
  share the first four steps but diverge at step five and receive opposite
  scores. These perturbations enter the multimodal feedback loop rather than
  remaining isolated pixel noise.}
  \label{fig:execution_instability}
\end{figure}

GUI execution is not reproducible at the level of byte-identical observations
or deterministic state transitions. Across clean resets, initial screenshots
already differ through clocks, tray state, cursor animation, notifications,
dynamic web content, antialiasing, and compositor timing. During execution,
rendering latency, focus, loading state, retained selections, and unrelated
dialogs can change the result of the same intended action. Because each new
screenshot conditions the next model decision and GUI actions target pixels,
small early differences can redirect subsequent actions and accumulate into a
different terminal state (Figure~\ref{fig:execution_instability}).

The endpoint can therefore flip for the same task, model, and harness. A failed
run alone does not establish that the task is unreachable, while a successful
run does not erase an informative failure path. A $K$-run task bundle exposes
both paths when they coexist, allowing the analyst to compare their divergence
instead of treating one noise-amplified branch as a stable mechanism. This
motivation agrees with recent computer-use reliability results showing that
outcome variation persists under deterministic decoding
\citep{gonzalezpumariega2026reliability}. Table~\ref{tab:fullset_flip} measures
its prevalence in the six model--harness configurations used by the optimizer.

\begin{table}[H]
  \centering
  \caption{Task-level flip rates under each baseline harness over $K=3$
  Full-361 executions. A task flips when its repeated runs contain both a
  zero score and a positive score.}
  \label{tab:fullset_flip}
  \small
  \setlength{\tabcolsep}{3.5pt}
  \begin{tabularx}{\textwidth}{l*{6}{>{\centering\arraybackslash}X}}
    \toprule
    & \makecell{Qwen3-VL-\\32B-Instruct}
    & \makecell{Qwen3-VL-\\8B-Instruct}
    & \makecell{Gemini\\3.1 Pro}
    & GPT-5
    & OpenCUA-72B
    & OpenCUA-32B \\
    \midrule
    Flip rate (\%) & 20.4 & 16.6 & 12.7 & 12.4 & 15.6 & 11.8 \\
    \bottomrule
  \end{tabularx}
\end{table}

Depending on the backbone, 11.8--20.4\% of tasks flip between zero and positive
score across three executions. The highest measured rates occur for
Qwen3-VL-32B-Instruct (20.4\%) and Qwen3-VL-8B-Instruct (16.6\%); Gemini 3.1 Pro
and GPT-5 still flip on 12.7\% and 12.4\% of tasks. Task-level bundles expose
both paths when they coexist and support comparisons at their divergence.

\subsection{Choosing the repeat count}

\begin{figure*}[t]
  \centering
  \includegraphics[width=\textwidth]{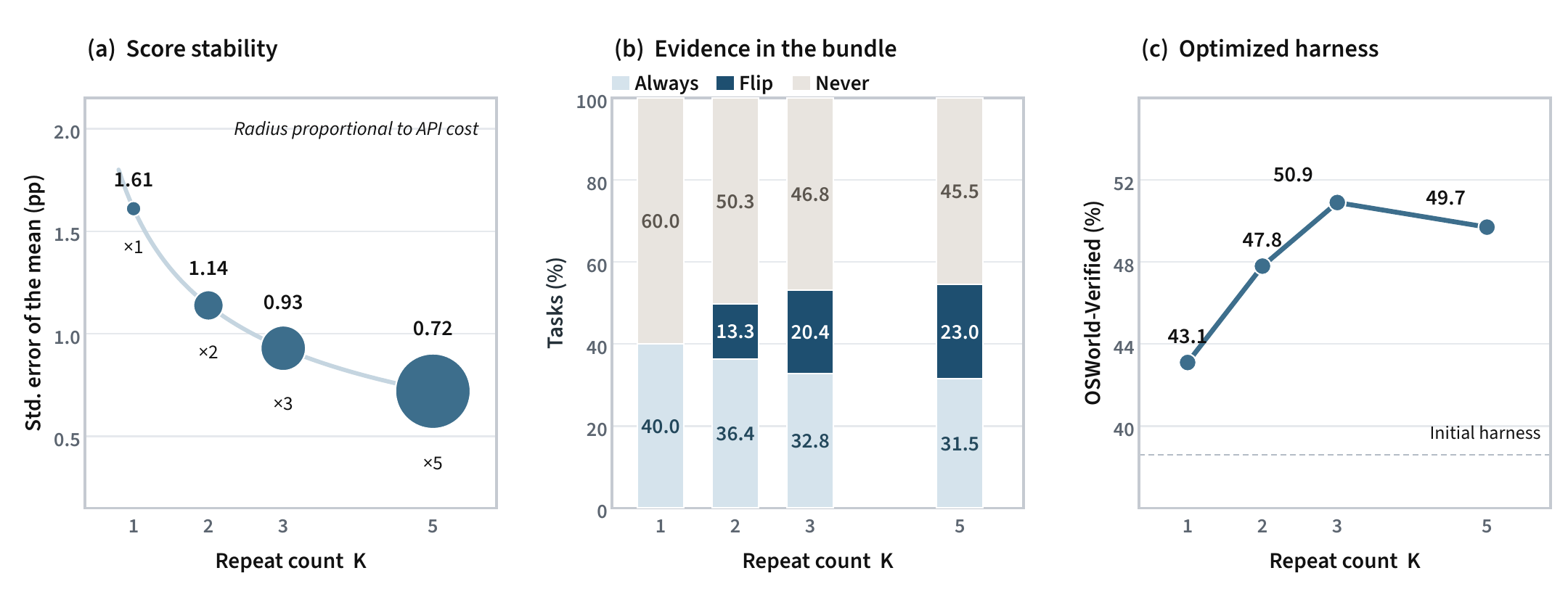}
  \caption{Effect of the repeat count $K$ for Qwen3-VL-32B-Instruct.
  (a) Standard error of the Full mean decreases with $K$, while rollout API
  cost scales approximately linearly. (b) Task-level flip rate as $K$ varies,
  with always-positive and never-positive shares shown for context. (c) Full
  score of the harness selected by each setting; the dashed line is the initial
  harness.}
  \label{fig:k_repeat_count}
\end{figure*}

The sweep separates evaluation precision, diagnostic coverage, and optimization
outcome. Increasing $K$ from 1 to 3 lowers the standard error from 1.61 to
0.93 points and reveals that 20.4\% of tasks contain both zero- and
positive-score outcomes. The selected harness improves from 43.1\% at $K=1$
to 50.9\% at $K=3$. Increasing to $K=5$ further lowers the standard error and
exposes more mixed-outcome tasks, but raises rollout cost to $5\times$ and
selects a slightly weaker 49.7\% harness. We therefore use $K=3$ throughout
the main experiments.

\section{Model-Specific Findings and Harness Evolution}
\label{app:dynamics}

\subsection{Behavior profiles across optimization}

The optimizer does not converge to one shared patch set. Table~\ref{tab:profiles}
connects representative task-level behavior changes to the executable runtime
mechanisms selected for each backbone. We report patterns that can be checked under
both the initial and final harness, while retaining overlapping counts when one
task exhibits more than one behavior.

\begin{table}[H]
\centering
\caption{Representative verified behavioral-pattern changes and the
corresponding runtime adaptations for each backbone. Counts denote Search-set
tasks matching each pattern under the initial and final harnesses
($H_0\!\rightarrow\!H^*$). Patterns are not mutually exclusive, so the counts
do not form a partition of failures.}
\label{tab:profiles}
\footnotesize
\setlength{\tabcolsep}{3.0pt}
\renewcommand{\arraystretch}{1.10}
\begin{tabularx}{\textwidth}{@{}l>{\raggedright\arraybackslash}X
  >{\raggedright\arraybackslash}p{3.55cm}
  >{\raggedright\arraybackslash}p{3.65cm}@{}}
\toprule
Backbone & Verified pattern changes ($H_0\!\rightarrow\!H^*$)
& Selected $H^*$ mechanisms & Residual bottleneck \\
\midrule
\rowcolor{gray!10}
\multicolumn{4}{@{}l}{\textit{General-purpose open models}} \\
Qwen3-VL-8B-Instruct
& redundant action loop $42\!\rightarrow\!27$; inefficient execution strategy
  $33\!\rightarrow\!19$; infeasibility misjudgment $10\!\rightarrow\!3$
& loop rejection and fresh restart; code routing; budget verdict; save nudge
& after a blocked action, the model may choose another ineffective route or
  emit \texttt{done()} again without correcting the state \\
\addlinespace[2pt]
Qwen3-VL-32B-Instruct
& inefficient execution strategy $36\!\rightarrow\!17$; premature completion
  $28\!\rightarrow\!15$; redundant action loop $11\!\rightarrow\!6$
& application-aware code routing; route ledger; save gate; completion
  adjudication; code receipt
& wrong targets or routes despite executed actions; incomplete work after code
  execution \\
\midrule
\rowcolor{gray!10}
\multicolumn{4}{@{}l}{\textit{GUI-specialized models}} \\
OpenCUA-32B
& zero-duration drag $33\!\rightarrow\!28$; unsupported mouse API
  $3\!\rightarrow\!0$; infeasibility misjudgment $3\!\rightarrow\!0$
& drag-duration normalization; \texttt{computer.*}-to-\texttt{pyautogui.*}
  mapping; infeasibility remapping
& correctly executed actions can still select the wrong target or drag
  distance; no alternative execution path is introduced \\
\addlinespace[2pt]
OpenCUA-72B
& unpersisted LibreOffice final state $24\!\rightarrow\!18$; dropped terminal
  action $4\!\rightarrow\!0$; infeasibility misjudgment
  $5\!\rightarrow\!3$
& pre-completion save; preservation of the last-step \texttt{DONE};
  text-grounded infeasibility handling
& ineffective or wrong GUI actions and visually plausible false completion \\
\midrule
\rowcolor{gray!10}
\multicolumn{4}{@{}l}{\textit{Proprietary frontier models}} \\
Gemini 3.1 Pro
& premature completion $20\!\rightarrow\!10$; tool--application state
  mismatch $13\!\rightarrow\!5$; inefficient execution strategy
  $11\!\rightarrow\!4$
& code routing; Office-file normalization; dry-run validation; constrained
  infeasibility criterion
& stale application views or semantically incorrect code output can still look
  complete \\
\addlinespace[2pt]
GPT-5
& inefficient execution strategy $27\!\rightarrow\!10$; redundant action loop
  $22\!\rightarrow\!10$; unnecessary post-tool continuation
  $15\!\rightarrow\!8$; infeasibility misjudgment $9\!\rightarrow\!2$
& code routing; Office persistence and restart; feasibility probe; code
  receipt; coordinate re-query
& semantic mismatch between a self-verified artifact and evaluator state;
  residual coordinate errors \\
\bottomrule
\end{tabularx}
\end{table}

Table~\ref{tab:profiles} retains the model-specific quantities and residual
bottlenecks. A complementary cross-backbone view in
Table~\ref{tab:crossmodel_modes} instead groups the same evidence by runtime
consequence, exposing shared structure without treating similar symptoms as
identical causes.

\begin{table}[H]
\centering
\caption{Cross-backbone organization of recurring, harness-addressable
behavioral patterns. The affected-backbone column lists models in which each
pattern was verified. A pattern groups model-specific manifestations with the
same runtime consequence; it does not imply an identical underlying cause.}
\label{tab:crossmodel_modes}
\footnotesize
\setlength{\tabcolsep}{3.0pt}
\renewcommand{\arraystretch}{1.10}
\begin{tabularx}{\textwidth}{@{}
  >{\raggedright\arraybackslash}p{3.05cm}
  >{\raggedright\arraybackslash}p{3.20cm}
  >{\raggedright\arraybackslash}X
  >{\raggedright\arraybackslash}p{4.10cm}@{}}
\toprule
Behavioral pattern & Affected backbones
& Shared intervention principle & Model-specific realizations \\
\midrule
\rowcolor{gray!10}
\multicolumn{4}{@{}l}{\textit{Execution and control}} \\
Inefficient execution strategy
& Qwen3-VL-8B and 32B; Gemini 3.1 Pro; GPT-5
& route bulk or repetitive operations through a shorter compatible execution
  path
& document maps and application references for Qwen; code routing and Office
  finalization for Gemini and GPT-5 \\
\addlinespace[2pt]
Redundant action loop
& Qwen3-VL-8B and 32B; GPT-5
& revise or withhold an action when repeated execution produces no useful state
  change
& loop rejection and fresh restart; route retirement; coordinate re-query and
  alternative code routing \\
\midrule
\rowcolor{gray!10}
\multicolumn{4}{@{}l}{\textit{Completion judgment}} \\
Premature completion
& Qwen3-VL-8B and 32B; OpenCUA-72B; Gemini 3.1 Pro
& require the task-relevant edit to be committed or persisted before accepting
  completion
& save nudge; save gate and completion adjudication; pre-completion save;
  Office-file normalization \\
\addlinespace[2pt]
Infeasibility misjudgment
& all six backbones
& use sufficiently supported infeasibility evidence to select the correct
  terminal outcome
& budget verdict and capability checks for general models; constrained
  feasibility probes for frontier models; text-grounded terminal remapping for
  OpenCUA \\
\midrule
\rowcolor{gray!10}
\multicolumn{4}{@{}l}{\textit{Tool and action interfaces}} \\
Tool--application state mismatch
& Qwen3-VL-32B; Gemini 3.1 Pro; GPT-5
& return structured execution evidence and synchronize the modified artifact
  with the visible application state
& code receipts; Office normalization and reload; controlled restart after
  external edits \\
\addlinespace[2pt]
Action-interface mismatch
& OpenCUA-32B and 72B; GPT-5
& normalize or preserve a valid model emission at the runtime dispatch boundary
& drag duration and API mapping; preservation of a valid terminal action;
  coordinate re-query \\
\bottomrule
\end{tabularx}
\end{table}

The labels describe observable behavior rather than assigning a unique causal
module. In particular, \emph{inefficient execution strategy} excludes tasks
that legitimately require GUI interaction or merely exceed a short step budget;
it applies when the evidence shows an avoidably low-throughput route despite an
available compatible alternative. This differs from a \emph{redundant action
loop}, where the same or equivalent action recurs without useful state progress.

Two recurring patterns illustrate why a shared label does not imply a universal
patch. Infeasibility misjudgment appears in all six backbones, but the selected
mechanisms range from budget and capability checks, to constrained feasibility
probes, to text-grounded terminal remapping. Likewise, premature completion
recurs across four backbones, yet is addressed through a save nudge, completion
adjudication, pre-completion persistence, or Office-file normalization according
to the available runtime interface.

The intervention granularity also separates along model families. Qwen3-VL-8B
relies on action-level enforcement when feedback alone does not change its next
action, whereas Qwen3-VL-32B combines enforcement with application-aware routing
and state evidence. The frontier models benefit mainly from routing,
persistence, and completion-policy changes. OpenCUA remains close to its native
coordinate-action interface, concentrating its adaptations on action
normalization and terminal semantics.

Across families, the remaining errors increasingly involve actions that execute
successfully but target the wrong object, route, or evaluator-relevant state.
Such trajectories provide less observable evidence for a runtime guard than an
unchanged screen, an invalid call, or an unsaved file, explaining why later
candidate edits are more likely to plateau or trade gains across tasks.

\subsection{Per-backbone harness evolution and final changes}

\begin{figure}[H]
  \centering
  \includegraphics[width=\textwidth]{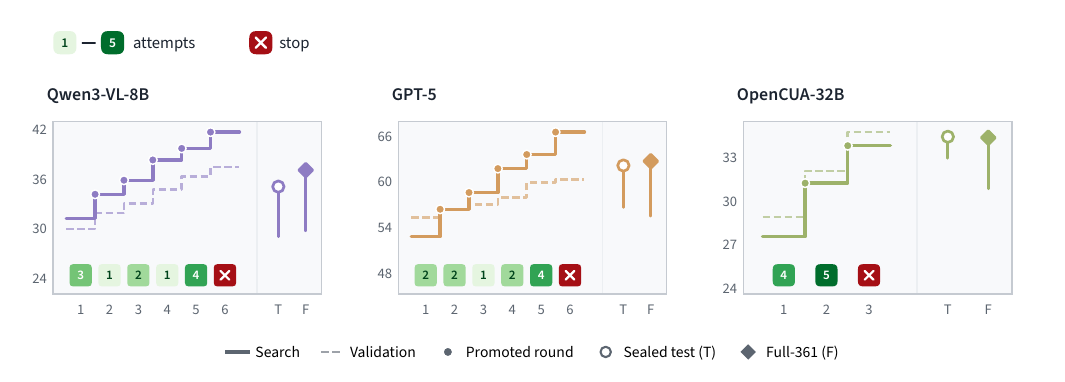}
  \caption{\textcolor{black}{Harness evolution for Qwen3-VL-8B, GPT-5, and
  OpenCUA-32B, complementing Figure~\ref{fig:harness_evolutions_main}.
  Solid and dashed curves
  show Search and Validation scores, respectively; filled circles mark promoted
  harnesses, and the rightmost markers show Test (T) and Full (F) scores after
  optimization stops. The numbered tiles report candidate-edit attempts in
  each optimization round, and the red cross marks the terminal round.}}
  \label{fig:harness_evolutions}
\end{figure}

\textcolor{black}{Figures~\ref{fig:harness_evolutions_main}
and~\ref{fig:harness_evolutions} expose} both accepted updates and the search
effort behind them. Qwen3-VL-32B promotes seven updates in eight rounds with 25
candidate-edit attempts; Qwen3-VL-8B promotes five in six rounds with 16
attempts. Gemini 3.1 Pro and GPT-5 promote six of seven and five of six rounds,
using 18 and 16 attempts, respectively. The specialized OpenCUA backbones stop
earlier: OpenCUA-72B promotes three updates in four rounds with 17 attempts,
whereas OpenCUA-32B promotes two in three rounds with 14 attempts. Thus the
optimization path depends on the backbone and initial harness: fewer rounds do
not necessarily imply fewer edit attempts or an easier adaptation problem.

\section{WindowsAgentArena Transfer}
\label{app:transfer}

\subsection{WindowsAgentArena benchmark context}

The paired transfer gains are summarized in the main paper.
Table~\ref{tab:waa} reports the corresponding runs, and
Table~\ref{tab:waa_benchmark} places them in the broader six-domain WAA context.
Published rows retain their reported runtimes, step budgets, and aggregation
protocol. The OSWorld optimization stage precedes WAA access; transfer supplies
only the platform-specific observation, action, and logging adapters.

\begin{table}[H]
\centering
\caption{Same-backbone transfer to WindowsAgentArena at 50 steps. The
GUI-HARVEST harnesses are selected on OSWorld and frozen before WAA evaluation;
no WAA trajectory enters optimization. The published Qwen operating point is
from \citet{yang2026ossymphony}.}
\label{tab:waa}
\small
\setlength{\tabcolsep}{8pt}
\renewcommand{\arraystretch}{1.10}
\begin{tabular}{llr}
\toprule
Backbone & Harness & WAA $\uparrow$ \\
\midrule
\multirow{3}{*}{Qwen3-VL-32B-Instruct}
& Qwen & 31.68 \\
& Agent S3 & 38.21 \\
& \textbf{\method} & \textbf{44.68} \\
\midrule
\multirow{2}{*}{GPT-5}
& Agent S3 & 50.88 \\
& \textbf{\method} & \textbf{64.75} \\
\bottomrule
\end{tabular}
\end{table}

\begin{table*}[t]
\centering
\caption{WindowsAgentArena results. Published rows retain their reported model,
runtime, and step budget; \method rows transfer the frozen OSWorld-derived
harness with platform adapters only. WAA contains six task domains
\citep{bonatti2025windows}. Published comparison values are from
\citet{yang2026ossymphony} and \citet{han2026vlaa}. For OS-Symphony rows,
Avg. is the published score over all 154 tasks, including its separately
reported 13-task infeasible subset; the six displayed columns are the WAA
application domains.}
\label{tab:waa_benchmark}
\small
\setlength{\tabcolsep}{4.2pt}
\resizebox{0.96\textwidth}{!}{%
\begin{tabular}{lrrrrrrrr}
\toprule
Method / backbone & Steps & Office & Web & System & Code & Media & Utility & Avg. \\
\midrule
\rowcolor{gray!16}\multicolumn{9}{c}{\itshape Max 50 Steps}\\
Qwen / Qwen3-VL-32B-Instruct~\citep{yang2026ossymphony} & 50 & 19.05 & 49.66 & 54.17 & 21.05 & 42.19 & 25.00 & 31.68 \\
OS-Symphony / Qwen3-VL-32B-Instruct~\citep{yang2026ossymphony} & 50 & 26.19 & 46.33 & 75.00 & 47.37 & 27.90 & 41.67 & 45.32 \\
VLAA-GUI / Gemini 3 Flash~\citep{han2026vlaa} & 50 & 32.60 & 73.30 & \textbf{87.50} & 66.70 & 52.40 & 75.00 & 60.40 \\
OS-Symphony / GPT-5-Mini~\citep{yang2026ossymphony} & 50 & 42.86 & 73.00 & 79.17 & 68.42 & 48.66 & 66.67 & 62.15 \\
OS-Symphony / GPT-5~\citep{yang2026ossymphony} & 50 & \textbf{54.76} & 73.00 & 75.00 & 42.11 & \textbf{70.09} & 75.00 & 63.45 \\
\midrule
\method / Qwen3-VL-32B-Instruct & 50 & 20.93 & 39.66 & 70.83 & \textbf{70.83} & 47.18 & 33.33 & 44.68 \\
\method / GPT-5 & 50 & 44.19 & 69.66 & \textbf{87.50} & 62.50 & 65.81 & \textbf{83.33} & \textbf{64.75} \\
\midrule
\rowcolor{gray!16}\multicolumn{9}{c}{\itshape Max 100 Steps}\\
VLAA-GUI / Gemini 3 Flash~\citep{han2026vlaa} & 100 & 35.00 & 73.30 & 87.50 & 66.70 & 52.40 & 83.30 & 61.00 \\
\bottomrule
\end{tabular}}
\end{table*}

\section{Cost and Efficiency}
\label{app:cost}

\subsection{Inference cost of the frozen harness}

\begin{figure}[H]
  \centering
  \includegraphics[width=\textwidth]{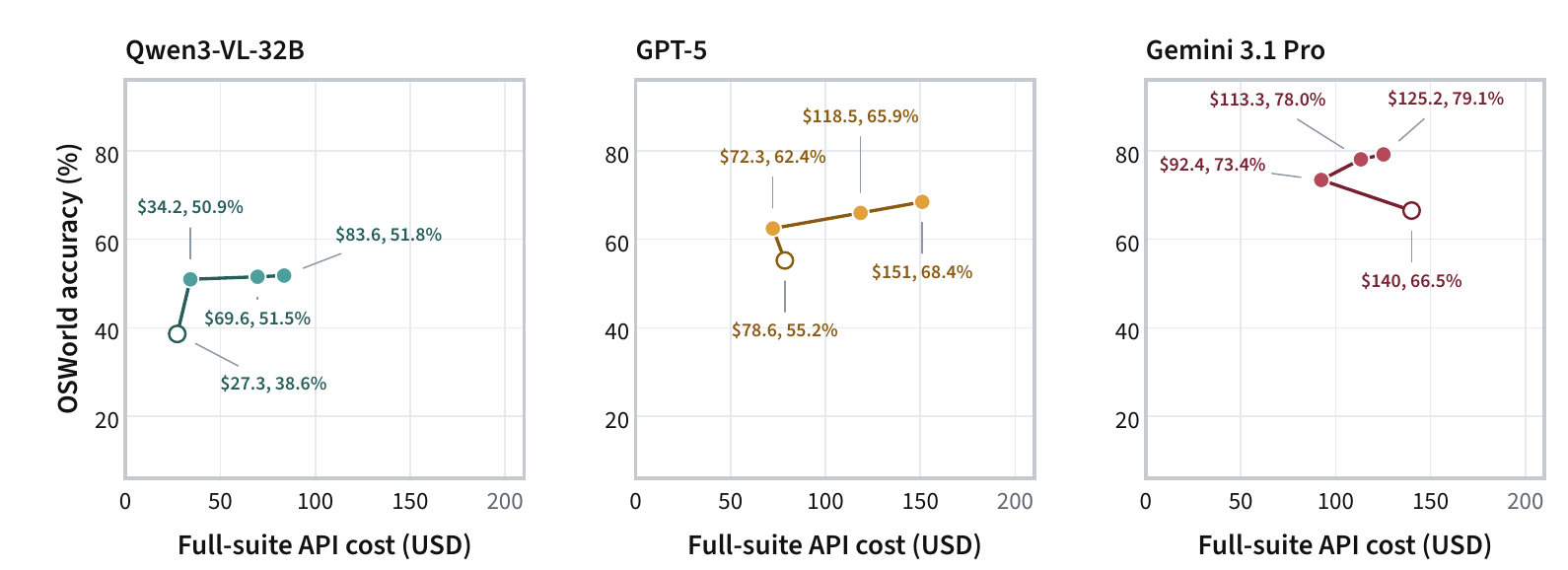}
  \caption{Full-suite API cost and OSWorld-Verified score for three target
  backbones. Open markers denote the 15-step Agent S3 baseline; filled markers
  denote the frozen GUI-HARVEST harness evaluated at 15, 50, and 100 steps.
  Costs cover target-model API calls for the 361-task evaluation.}
  \label{fig:paired_cost}
\end{figure}

Figure~\ref{fig:paired_cost} shows that harness adaptation changes both task
success and the number and length of model calls. For Qwen3-VL-32B-Instruct,
the 15-step optimized harness raises accuracy from 38.6\% to 50.9\% while cost
increases from \$27.3 to \$34.2. For GPT-5 and Gemini 3.1 Pro, the optimized
15-step harness is both more accurate and less expensive: GPT-5 moves from
55.2\% at \$78.6 to 62.4\% at \$72.3, and Gemini moves from 66.5\% at \$140.0
to 73.4\% at \$92.4. These are end-to-end measurements of the frozen
harnesses.

The published comparison in Figure~\ref{fig:benchmark_cost} uses cost--accuracy
records from \citet{wei2026steplevel} and
\citet{gonzalezpumariega2025scaling}. The former directly reports the following
accuracy and cost/task pairs: 43.3\%/\$0.022 for EvoCUA-8B,
30.8\%/\$0.018 for Qwen3-VL-8B, 55.4\%/\$0.224 for EvoCUA-8B plus
Claude Sonnet 4.5, 54.3\%/\$0.423 for Qwen3-VL-8B plus Claude Sonnet 4.5,
and 58.1\%/\$0.881 for Claude Sonnet 4.5. Its switched-task counts and
percentages imply 359 evaluated tasks (e.g., $168/46.8\%$); the plotted totals
multiply the reported per-task costs by 359. The Agent S3 point combines
the 62.6\% 100-step accuracy and \$0.72 average cost/task reported by
\citet{gonzalezpumariega2025scaling}; \$260 is $361\times\$0.72$, rounded.
Thus, the plotted public totals are direct arithmetic reconstructions from
source-reported records rather than independent token-cost estimates.

For our runs, token usage is converted to API cost using the official provider
prices in effect during evaluation: OpenAI and Google prices for the
proprietary models, and Alibaba Cloud Model Studio prices for Qwen3-VL.
Optimization calls and local compute are excluded. Public points retain the
source paper's pricing assumptions. Systems without sufficient usage or cost
information are omitted rather than assigned estimated values.

\section{More Implementation Details}
\label{app:implementation}

\subsection{Model usage and inference settings}

The target backbones are Qwen3-VL-8B/32B-Instruct, OpenCUA-32B/72B,
Gemini 3.1 Pro, and GPT-5. GPT-5 uses the dated
\texttt{gpt-5-2025-08-07} endpoint. For Qwen3-VL and the proprietary
backbones, Agent S3 uses its default UI-TARS-1.5-7B grounding model in the
native $1920\times1080$ coordinate space; this grounder is part of the initial
harness rather than an additional optimized model. OpenCUA uses its official
coordinate-action runtime. All four optimizer roles use Claude Sonnet 5 with
identical role prompts across target backbones. Target-model weights, API
snapshots, decoding parameters, observation format, and maximum output length
remain fixed within each experiment.

\subsection{Environment and benchmark details}

OSWorld-Verified is executed in clean virtual-machine snapshots with the
benchmark's programmatic evaluators \citep{xie2024osworld}. Following common
practice, the main protocol excludes the eight Google Drive tasks and uses the
remaining 361 tasks \citep{han2026vlaa}. The execution configuration fixes the
OSWorld commit, VM and application images, screen resolution, action limit,
timeout, reset and retry policy, task-group manifest, and evaluator versions.
Search, validation, and test groups are fixed before optimization. The final
full-361 run follows the frozen harness and stopping decision.

\textcolor{black}{WindowsAgentArena evaluates frozen-harness transfer for
Qwen3-VL-32B-Instruct and GPT-5 at a 50-step budget
\citep{bonatti2025windows}. OSWorld optimization completes before WAA access.
Following the transfer setups of Agent S3 and OS-Symphony
\citep{gonzalezpumariega2025scaling,yang2026ossymphony}, the implementation
maps Windows observations, actions, and logging interfaces and updates
platform-specific prompt and command conventions, including Linux-to-Windows
application and shell wording. These are necessary benchmark-interface
adaptations: the learned runtime mechanisms remain fixed, and no WAA trajectory
enters optimization or harness selection.}

\subsection{Initial harnesses and action contracts}

\paragraph{Agent S3.}
Agent S3 is the initial harness for Qwen3-VL and the proprietary backbones. It
is an open, competitive OSWorld runtime and a widely used reference baseline,
with an explicit worker, reflection, grounding, and optional code-execution structure
\textcolor{black}{\citep{gonzalezpumariega2025scaling,qin2025uitars}}. Its worker emits exactly one
\texttt{agent.*} primitive per step. Pointing primitives carry a natural-language
description of the target; the default UI-TARS-1.5-7B grounder maps that
description to coordinates and Agent S3 compiles the result into executable
PyAutoGUI code. The grounder is part of the initial harness. We retain all
default runtime settings except the target backbone. Behavior best-of-$N$ is
disabled ($N=1$), so each task attempt contributes one trajectory.

\paragraph{Official OpenCUA runtime.}
OpenCUA is trained with an action space that directly represents keyboard and
coordinate-bearing mouse operations as PyAutoGUI actions
\citep{wang2025opencua}. OpenCUA-32B/72B begin from the official OpenCUA agent.
Agent S3 expects a different \texttt{agent.*} syntax followed by a separate
grounding stage, whereas OpenCUA post-training directly learns
coordinate-bearing PyAutoGUI actions. For these rows, $H_0$ denotes the official
OpenCUA runtime and GUI-HARVEST edits its corresponding executable surface.

\subsection{\textcolor{black}{Optimization and ablation protocol}}

\textcolor{black}{Search uses a 15-step environment budget. Unless identified
as an externally published comparison, every reported GUI-HARVEST score first
averages $K=3$ independent runs within each task and then averages across tasks.
Search, Validation, Test, and Full scores all follow this protocol, consistent
with the repeated-run
protocol used in recent computer-use reliability analysis
\citep{gonzalezpumariega2026reliability}. Appendix~\ref{app:repeated} examines
the trade-off between repeat count, diagnostic coverage, and rollout cost.
After optimization, the same frozen harness is evaluated at 15, 50, and 100
steps without further edits. The task split, evaluator, repeat count, round
cap, and promotion rule are shared across backbone families.}

\textcolor{black}{The main protocol allows at most $B=10$ rounds per backbone.
Each round searches from one fixed harness state, and up to $P=5$ L1-valid
candidates may enter GUI evaluation. Promotion starts the next round with a
fresh five-attempt budget;
if all five candidates are rejected, optimization stops at the current harness.
L0/L1 repairs made before a valid candidate enters GUI evaluation consume no
attempt. All stopping
parameters are fixed before the sealed test is opened. A capability manifest
permits edits to prompts, context and memory
construction, action interfaces, control flow, verification, recovery, and
termination, while protecting the model client, benchmark, evaluator, tasks,
and resource limits.}

\textcolor{black}{Component ablations use Qwen3-VL-32B-Instruct and remove one
element at a time: visual evidence, the Cross-task Clusterer, repeated
optimizer evidence ($K=1$), or the behavioral soft gate. Variants share the
initial harness, editable scope, task split, and optimization-round cap.
All selected harnesses, including the $K=1$ optimizer variant, use the same
$K=3$ final evaluation protocol (Table~\ref{tab:ablation_joint}).}

\subsection{Per-backbone optimization records}

Figure~\ref{fig:harness_evolutions} summarizes promoted rounds and candidate
edit attempts for all six target backbones. Tables~\ref{tab:profiles}
and~\ref{tab:crossmodel_modes} connect the retained behavior changes to the
runtime mechanisms in each final harness.

\subsection{Role interfaces, prompts, and schemas}

Evidence analysts and intervention validators are short-lived task-local
workers that run in parallel. The Cross-task Clusterer receives verified
task-local findings grouped by outcome category and induces recurring behavior
modes supported by independent tasks. The harness engineer
is one bounded tool session per candidate and can search, edit, and test only
paths allowed by the capability manifest.
Deterministic programs perform log parsing, citation checks, score aggregation,
L0/L1 checks, promotion, rollback, and ledger writes.

\begin{figure}[t]
  \centering
  \includegraphics[width=0.92\textwidth]{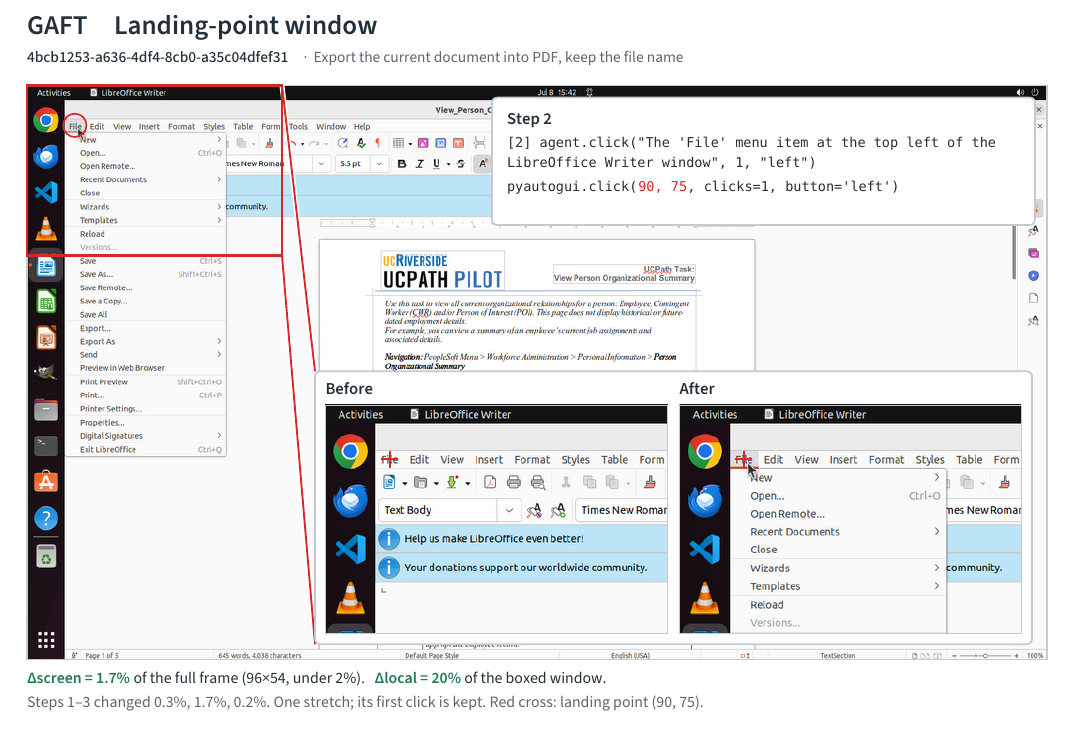}
  \caption{GAFT evidence for a pointing action. The toolkit aligns the executed
  click with before/after screenshots and reports pixel change over the full
  screen and a local window around the landing point. These measurements are
  mechanical evidence supplied to the Evidence Analyst, not semantic judgments.}
  \label{fig:gaft_landing_callout}
\end{figure}

GAFT converts each raw execution into a step-aligned evidence bundle. For every
step it retains the model text, planned and executed action, tool outputs, and
before/after screen references, and computes only reproducible visual facts:
the whole-screen changed-pixel fraction $\Delta_{\mathrm{screen}}$ and, for
pointing actions, $\Delta_{\mathrm{local}}$ in a fixed window around the landing
point. The initial analyst context includes event-selected keyframes---the
initial and final screens, large screen transitions, and the start of sustained
no-change stretches---plus landing-point crops when they are informative. The
analyst may then request additional full-resolution frames or crops by run and
step. Retrieval expands the evidence available for semantic diagnosis without
asking the preprocessing code to infer what the screen means.
Figure~\ref{fig:gaft_landing_callout} illustrates the landing-point evidence
exposed for a pointing action.

A verified finding identifies the failed run and decisive step, summarizes the
observed action--state mismatch, and links it to run/step evidence; when
available, it also records a successful-run contrast. Screens are supplied with
the task bundle and can be retrieved by run and step during analysis. A behavior mode contains an outcome category,
observable mechanism, membership test, behavioral target, member findings, and
independent-task support. Findings that do not form a recurring mode remain
available as residue; the engineer may consider one only when it can justify a
shared source lever and a cross-task-safe intervention. The update manifest
connects selected targets and source locations to the patch, target tasks,
predicted changes, and protected interfaces. Validator outputs contain one
task-level verdict and run/step citations.

Appendix~\ref{app:evidence_chain} presents an end-to-end record exported from
the implementation.

\subsection{Split and leakage controls}

\textcolor{black}{Tasks are assigned to disjoint Search, Validation, and Test
sets of 80, 80, and 201 tasks through a domain-stratified random split. Each
domain contributes in proportion to its size; tasks within a domain are sampled
uniformly with a public seed. The assignment is fixed before rollout and reused
across backbones.}
The optimizer receives Search task cards and rollout artifacts. Validation execution is
performed by the outer runner, which exports the aggregate mean needed by
the score gate. Test task identifiers and outcomes remain inaccessible until
the selected harness commit is frozen. The fixed task manifest records the
group identifier for every task.

\subsection{Comparison methods}
\label{app:optimizer_comparison}

\begingroup
\color{black}
\paragraph{Paired harness comparisons.}
The $H_0$--$H^*$ comparisons fix the target backbone and evaluation protocol
to measure the effect of harness adaptation. Full scores additionally
provide operating points for comparison with published systems; Test
scores measure generalization to tasks withheld during optimization.

\paragraph{Harness optimizers.}
For Self-Harness and Meta-Harness, we freeze Qwen3-VL-32B-Instruct and start
from the same Agent S3 commit and writable source tree used by \method. Both
adaptations use Claude Sonnet 5 and a maximum of ten optimization rounds.
Self-Harness receives serialized Agent S3 plans, actions, tool outputs, and
scores through its trajectory interface; Search and Validation serve as its
held-in and held-out sets. We retain its failure-key clustering, single-hook
proposals, and score-based acceptance rule. Meta-Harness receives a filesystem
containing candidate source, Search scores, trajectories, instructions,
results, and screenshots. It proposes two harnesses per round and selects the
final harness by Search score, with Validation and sealed Test withheld from
selection. These adapters change data formats and GUI execution plumbing, not
the optimizers' diagnosis or selection policies.

The LFF paper and repository do not provide a complete experimental
configuration, an executable optimization procedure, or the optimizer/framework
source required to rerun the search. The repository provides the resulting
OpenCUA patch artifacts. We therefore apply those released patches to the
matching OpenCUA-32B and OpenCUA-72B runtimes and evaluate them on our 15-step
Test and Full protocol. Table~\ref{tab:optimizer_comparison}
therefore reports within-backbone comparisons and does not aggregate across the
Qwen and OpenCUA blocks.

\paragraph{Published GUI systems.}
Comparisons include Agent S3, OS-Symphony, VLAA-GUI, OpenCUA, LFF, and related
systems. Each published row retains its reported model, runtime, step budget,
and aggregation protocol (Appendix~\ref{app:additional}). LFF's 100-step
OpenCUA-32B/72B results are retained as published operating points in
Table~\ref{tab:benchmark_opencua} \citep{sun2026learning}; they come from
independent executions and are not paired trials against our initial harnesses.
\par
\endgroup

\subsection{Runtime validity and failure handling}

Each rollout starts from the benchmark's clean snapshot. Runs with environment
startup failure, missing screenshots, evaluator failure, or unrecoverable runner
exceptions are marked invalid and rerun under a predeclared retry policy. The
policy is identical for the current and candidate harness. Partial evaluator
scores are retained as raw values. A bundle enters failure diagnosis when at
least one valid run has score zero. Zero-score runs are finding targets, while
the remaining runs provide comparison evidence.

\subsection{\textcolor{black}{Evaluation metrics and cost scope}}

\textcolor{black}{The primary metric is the mean raw evaluator score, averaging
the $K$ runs within each task and then across tasks (Eq.~\ref{eq:utility}).
Scores are expressed as percentages, and $H_0$--$H^*$ gains are absolute
percentage-point differences. For the repeated-execution analysis in
Appendix~\ref{app:repeated}, a task is \emph{always positive} if all $K$ runs
have positive scores, \emph{never positive} if none does, and a \emph{flip}
if both zero- and positive-score outcomes occur. This decomposition captures
outcome variation that is hidden by a single rollout.}

\textcolor{black}{Full-suite inference API costs cover target-model calls during
the 361-task evaluation of a frozen harness. Optimization-process costs are
not included in these measurements. Appendix~\ref{app:cost} specifies the
scope and pricing caveats for published cost comparisons.}

\section{Formal Definitions for GUI-HARVEST}
\label{app:formal_method}

This appendix \textcolor{black}{summarizes the optimization loop and} specifies the records and promotion rule used in
Section~\ref{sec:method}. Utility is defined in Equation~\ref{eq:utility}, and
the score-based acceptance condition is given in Equation~\ref{eq:hardgate}.

\begin{algorithm}[H]
\color{black}
\caption{\textcolor{black}{GUI-HARVEST optimization}}
\label{alg:harvest}
\begin{algorithmic}[1]
\Require Frozen model $M$, seed harness $H_0$, splits
$\searchset,\valset,\testset$, repeats $K$, round cap $B$, attempts per round $P$
\State $H\gets H_0$; $\ledger\gets\emptyset$; evaluate $H$ on
$\searchset,\valset$ with $K$ rollouts per task
\For{$b=1,\ldots,B$}
    \State Build or reuse search bundles from rollouts of the current $H$
    \State Evidence Analyst: diagnose failed runs with verified evidence, or
    reuse existing findings
    \State Cross-task Clusterer: induce recurring modes; reuse verified modes
    if $H$ is unchanged
    \State $\mathit{promoted}\gets\mathrm{false}$
    \For{$p=1,\ldots,P$}
        \State Harness Engineer: propose $\widetilde H$, target tasks, and
        predicted changes; repair until L0/L1-valid
        \State Validator: evaluate $\widetilde H$ on $\searchset,\valset$ with
        $K$ rollouts per task; apply utility and behavior gates
        \State Append the patch, predictions, scores, and verdicts to $\ledger$
        \If{$\widetilde H$ is promoted}
            \State $H\gets\widetilde H$; retain its search rollouts
            \State $\mathit{promoted}\gets\mathrm{true}$; \textbf{break}
        \Else
            \State Roll back to $H$
        \EndIf
    \EndFor
    \If{$\mathit{promoted}=\mathrm{false}$} \State \textbf{break} \EndIf
\EndFor
\State Freeze $H^*\gets H$; perform the sealed-test evaluation on $\testset$
\State \Return $H^*$
\end{algorithmic}
\end{algorithm}

\subsection{Execution evidence and findings}
\label{app:formal_evidence}

For task $x_i$, run $r$ under the current harness $H_t$ at round $t$ is recorded
as
\begin{equation}
\label{eq:run_record}
    \rho_{i,r}^{(t)}=
    \left(\tau_{i,r}^{(t)},s_{i,r}^{(t)},z_{i,r}^{(t)}\right),
\end{equation}
where $\tau_{i,r}^{(t)}$ is the multimodal trajectory, containing model text,
planned and executed actions, screenshots, and tool outputs;
$s_{i,r}^{(t)}$ is the run's benchmark score; and $z_{i,r}^{(t)}$ contains runtime
and termination metadata. Repeated runs are grouped into a task-round bundle:
\begin{equation}
\label{eq:bundle}
    \bundle_i^{(t)}=
    \left(x_i,\mathcal C(H_t),\{\rho_{i,r}^{(t)}\}_{r=1}^{K}\right).
\end{equation}
The task record $x_i$ contains the instruction, domain, related applications,
feasibility annotation, setup steps, and any hint published with the task. The
Harness Card $\mathcal C(H_t)$ summarizes the current runtime and its observable
outputs. The bundle separates this shared information from the evidence
specific to each run.

The Evidence Analyst produces findings of the form
\begin{equation}
\label{eq:finding}
    f=(r,\,t^\star,\,\sigma,\,m,\,\bar a,\,E),
\end{equation}
where $r$ identifies the failed run, $t^\star$ is its decisive step,
$\sigma$ records the decisive action, observed outcome, and subsequent response,
$m$ is the observable mechanism, $\bar a$ is an evidence-supported alternative
behavior when available, and $E$ contains run/step evidence and any successful-run
contrast. Findings remain linked to their task bundle and harness version. The
evidence checks in Section~\ref{sec:evidence} validate these records before
clustering.

\subsection{Behavior modes and update plans}
\label{app:formal_updates}

The Cross-task Clusterer first maps each finding signature $\sigma$ to a
deterministic outcome category $g$. It then represents each recurring behavior
mode within one category as
\begin{equation}
\label{eq:mode}
    c=(g,\,d_c,\,\phi_c,\,b_c,\,F_c),
\end{equation}
where $d_c$ describes its observable mechanism, $\phi_c$ is a membership test,
$b_c$ is the shared behavioral target, and $F_c$ contains member findings from
at least two independent tasks. Category $g$ narrows the clustering search
space without fixing the mode vocabulary or the eventual intervention. Member
findings retain links to their task-local evidence.

Before candidate evaluation, the Harness Engineer records an update plan
\begin{equation}
\label{eq:update}
    u=(c,E,a,\delta,\mathcal{P},A,q).
\end{equation}
Here $c$ and $E$ identify the target mode and evidence, $a$ is the source surface
being changed, $\delta$ is the intervention, $\mathcal{P}$ contains observable
predictions, and $A\subseteq\searchset$ contains the target tasks used to check
them. The invariants $q$ specify interfaces or behaviors that must remain
intact. The plan links the proposed code change to evidence that can be checked
in subsequent executions.

\subsection{Candidate scoring and promotion}
\label{app:formal_promotion}

For each split $D\in\{\searchset,\valset\}$, the candidate's score change is
\begin{equation}
\label{eq:score_change}
    \Delta_D=\widehat J_{D,K}(\widetilde H_t)-
    \widehat J_{D,K}(H_t).
\end{equation}
The utility hard gate in Equation~\ref{eq:hardgate} requires both changes to be
nonnegative and at least one to be strictly positive. Validation contributes
only aggregate scores; detailed execution evidence is available only for
search tasks.

The behavioral soft gate checks every prediction in $\mathcal{P}$ against the
post-edit executions of its target tasks in $A$. Each prediction requires at
least one determinate task verdict (supported or contradicted) and more
supporting than contradicting verdicts. L0 and L1 denote the code checks
defined in Section~\ref{sec:validation}. Subject to the fixed execution-cost
constraint, the candidate is promoted exactly when
\begin{equation}
\label{eq:promote}
    \operatorname{Promote}(\widetilde H_t)=
    \mathrm{L0}\land\mathrm{L1}\land
    \operatorname{HardGate}_{\searchset,\valset}\land
    \operatorname{SoftGate}_{\searchset}.
\end{equation}
Promotion sets $H_{t+1}=\widetilde H_t$; rejection keeps the current harness
unchanged. The update ledger retains the patch, predictions, score changes,
task verdicts, and decision for subsequent rounds.

\section{A Complete Evidence-to-Edit Chain}
\label{app:evidence_chain}

Figure~\ref{fig:evidence_chain} reproduces one end-to-end record from the
optimizer. The task asks the agent to rename a Chrome profile to Thomas. In all
three pre-edit executions, the field visibly contains the requested name; two
runs nevertheless call \texttt{done()} while the field remains in edit mode and
fail, whereas the successful run first clicks outside the field. The analyst
therefore identifies an uncommitted edit rather than equating visible text with
a completed state change. The clusterer links this finding to independent
Impress and Calc tasks in which the requested edit is likewise visible but
unsaved when termination is requested, yielding the cross-task mode
\textsc{Edit-Not-Committed-at-Done}. The engineer implements an action gate that
tracks edit actions, recognizes explicit saves, and withholds the first
premature \texttt{done()} while returning the objection to the planner.
It also records the prediction that a withheld termination should be followed by
a commit action before a later \texttt{done()} is accepted. The validator then
checks three post-edit executions of the Chrome task. Each first termination is
withheld; the agent leaves the field, and termination is accepted only after the
committed name is observable again. This chain shows how repeated executions
separate a stable behavioral defect from a successful contrast, while visual
state distinguishes typing the requested value from completing the edit.

\noindent\makebox[\textwidth][c]{%
  \includegraphics[width=0.88\textwidth]{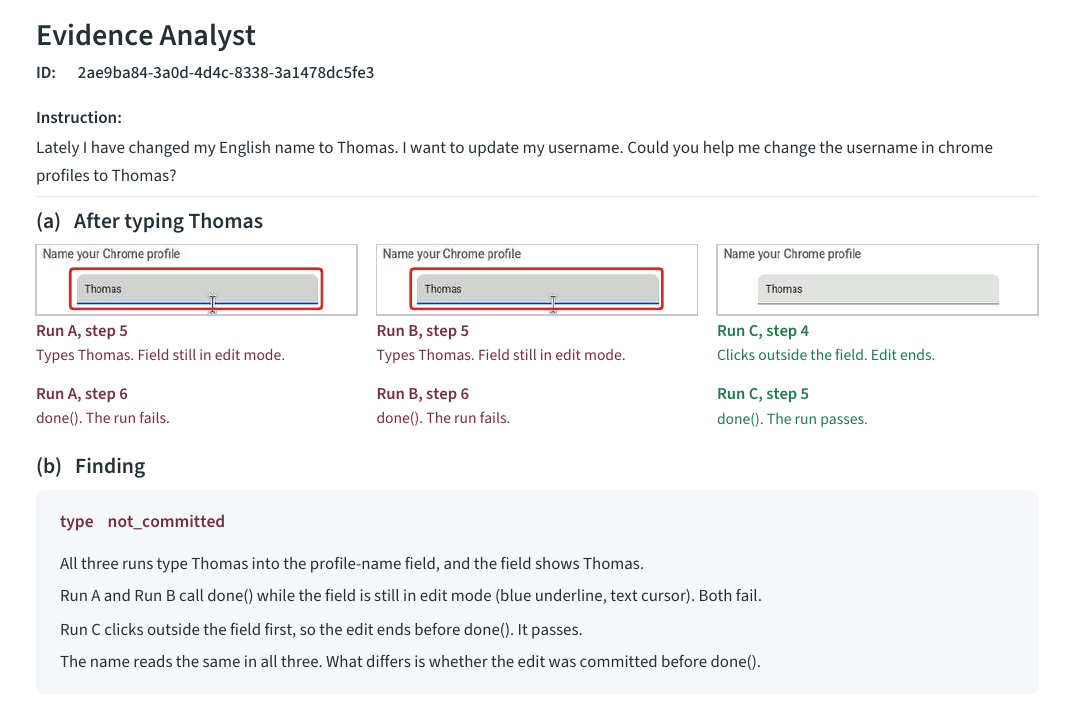}}\par\vspace{1mm}
\noindent\makebox[\textwidth][c]{%
  \includegraphics[width=0.88\textwidth]{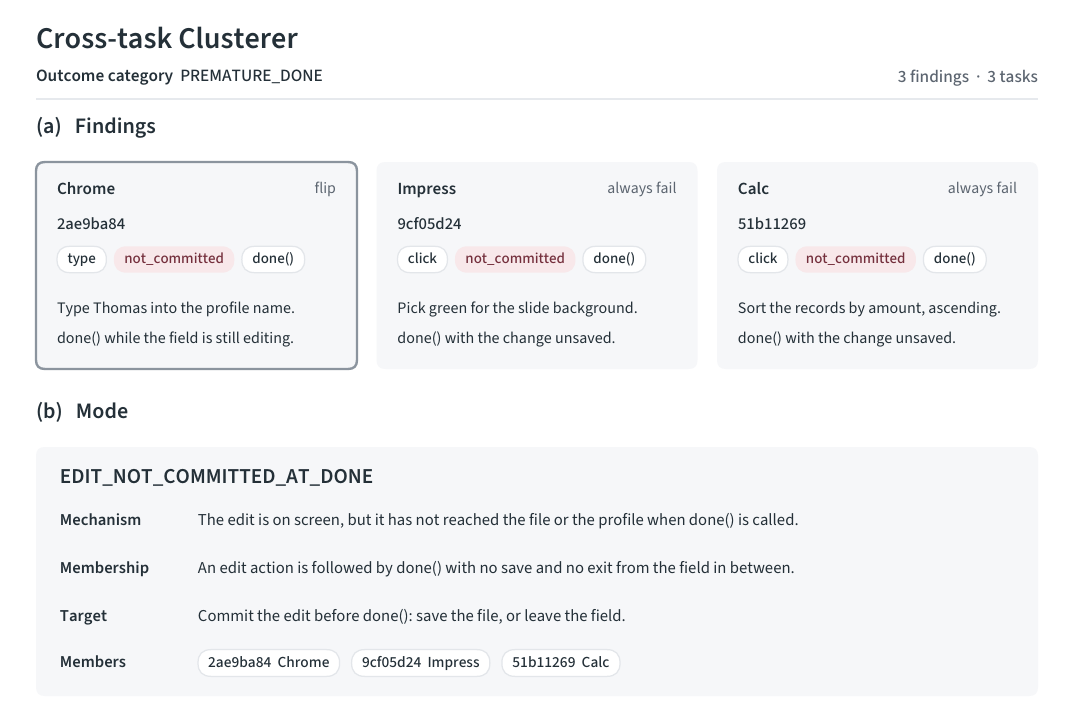}}\par\vspace{1mm}
\noindent\makebox[\textwidth][c]{%
  \includegraphics[width=0.88\textwidth]{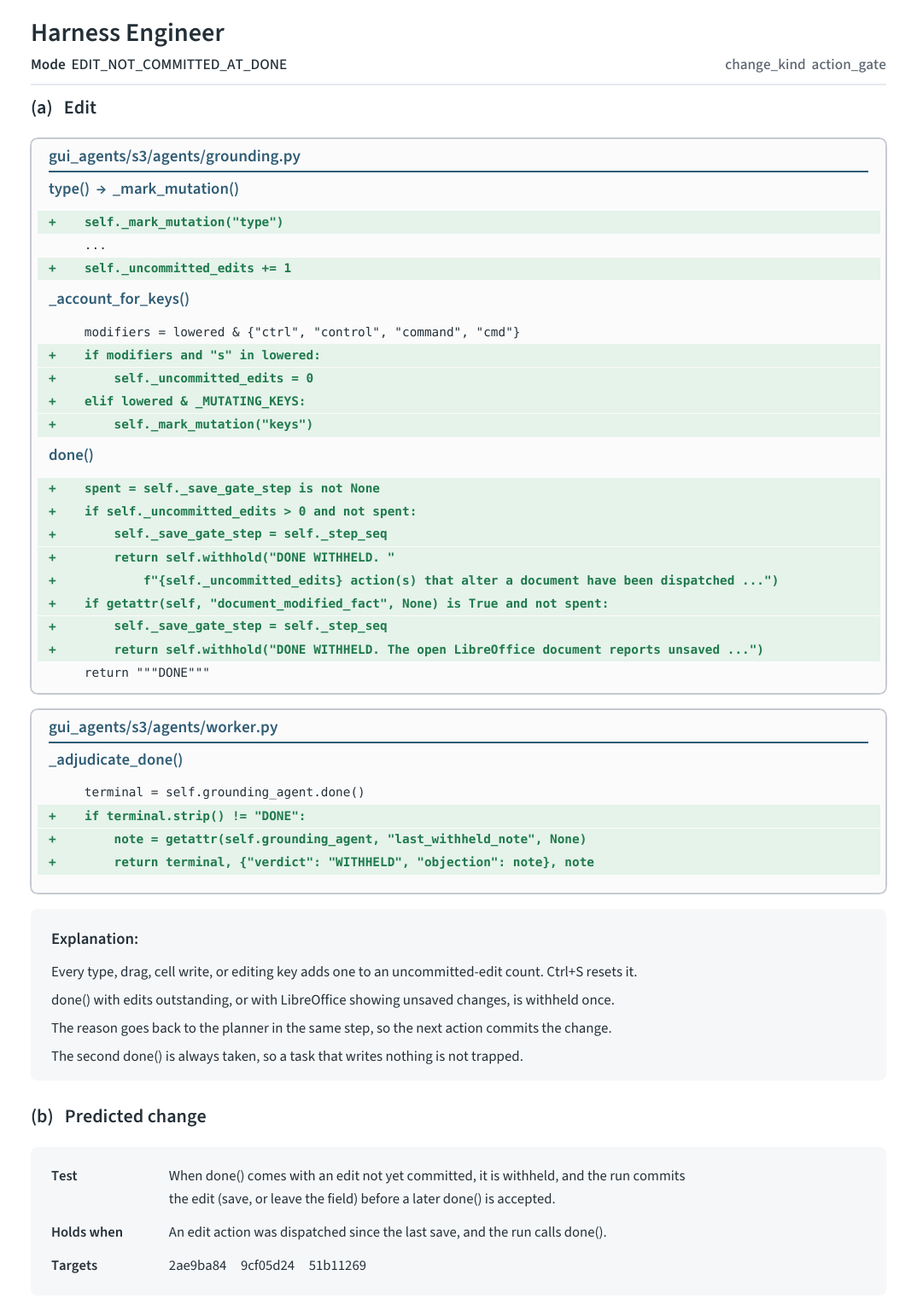}}\par\vspace{1mm}
\noindent\makebox[\textwidth][c]{%
  \includegraphics[width=0.88\textwidth]{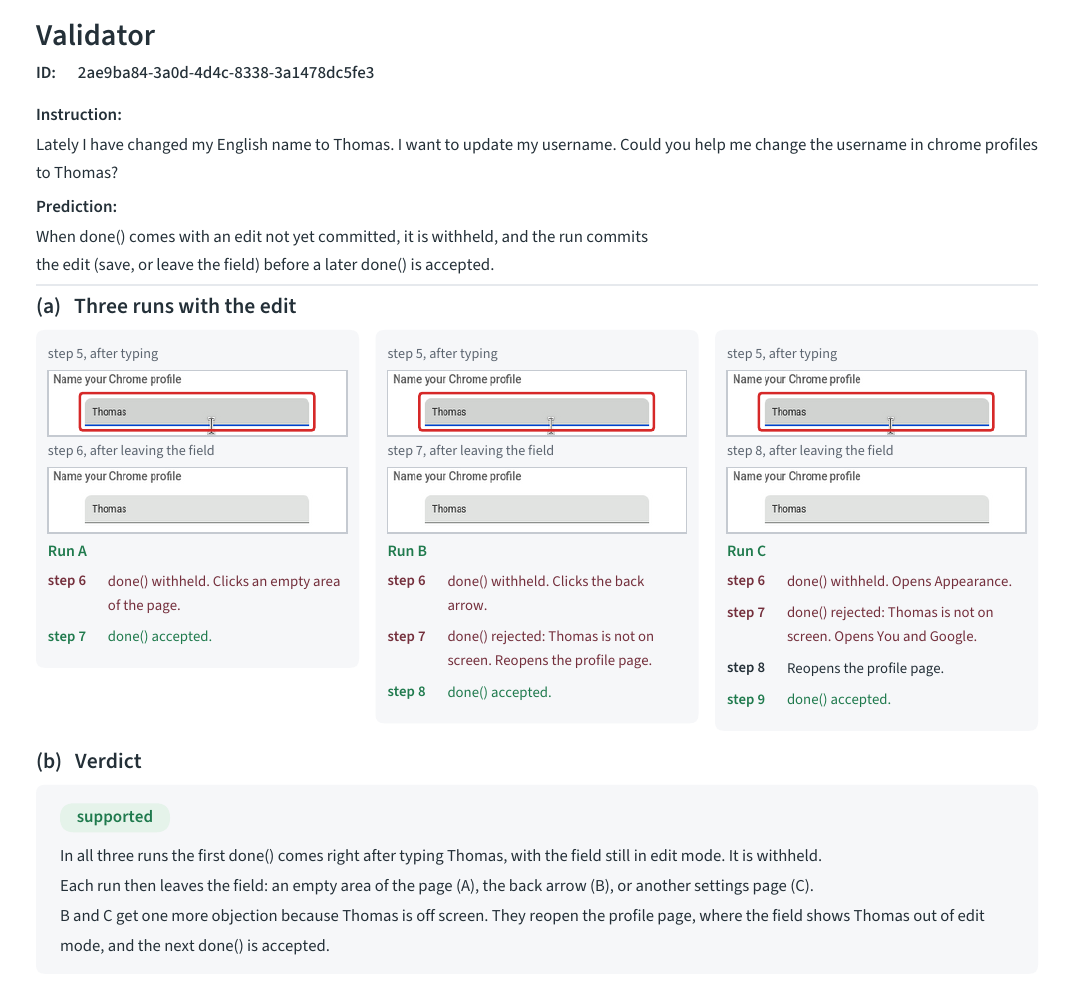}}\par

\captionof{figure}{Real end-to-end evidence chain from repeated multimodal
executions to an executable harness update. Two failed Chrome runs terminate
with the renamed profile field still in edit mode, while a successful contrast
commits the edit first. The finding is aggregated with analogous Impress and
Calc failures, translated into a termination gate, and supported by three
post-edit executions. Task, run, and step links are retained throughout
diagnosis, clustering, editing, and behavioral validation; promotion additionally
requires the Search/Validation utility gate.}
\label{fig:evidence_chain}

\section{Representative Role Prompt Templates}
\label{app:prompts}

The following templates are abridged from the prompts used in our experiments.
They preserve the model-visible inputs, decision rules, and structured output
contracts while eliding long vocabularies and repeated formatting checks.

\newtcolorbox{prompttemplate}[1]{
  enhanced,
  breakable,
  colback=black!3,
  colframe=black!72,
  colbacktitle=black!72,
  coltitle=white,
  title={#1},
  fonttitle=\bfseries\small,
  arc=1.6mm,
  outer arc=1.6mm,
  boxrule=0.65pt,
  left=2.5mm,
  right=2.5mm,
  top=1.6mm,
  bottom=1.6mm,
  before skip=6pt,
  after skip=9pt,
  titlerule=0pt,
  toptitle=1.2mm,
  bottomtitle=1.2mm
}

\begin{prompttemplate}{Evidence Analyst \;|\; Task-round diagnosis}
\begin{Verbatim}[fontsize=\scriptsize]
You are given every run of ONE task in the current round --
{n_runs} runs, each with its benchmark score.
At least one scored 0.

YOUR QUESTION
For each run that scored 0: at which step did it stop getting
nearer to the end state required by the task, and what exactly did it
do there? Answer in observations, not in causes.

WHAT YOU ARE GIVEN
(1) HARNESS CARD: loop, primitives, semantics, recorded fields.
(2) TASK CARD: instruction, domain, related apps, feasibility,
    setup steps, and any task-published hint.
(3) RUNS: actions, planner text, screenshots, and run scores.

PROCEDURE
1. If some run scored, use it as the REFERENCE and find the
   first outcome-relevant divergence.
2. Find the earliest DECISIVE STEP that introduced an error
   not fully repaired later.
3. Record did, machine, signature, mechanism, counterfactual.
4. Ground the finding in a verbatim decisive-step quote.

OUTPUT (one JSON object)
{
  "screens_wanted": [{"run":"C","step":7,"kind":"before"}],
  "findings": [{
    "run":"C", "decisive_step":N,
    "contrast":{"reference_run":"A","step":N,
                "reference_did":"..."} | null,
    "signature":{"outcome":"...","noticed":"..."},
    "did":"...", "machine":"...", "mechanism":"...",
    "counterfactual":"..." | null, "quotes":["VERBATIM"]
  }]
}
\end{Verbatim}
\end{prompttemplate}

\begin{prompttemplate}{Cross-task Clusterer \;|\; Recurring behavior modes}
\begin{Verbatim}[fontsize=\scriptsize]
Outcome category {outcome_category}: {outcome_category_def}.

Below are all {n} findings in this outcome category from
{n_tasks} tasks. Each contains an Analyst mechanism, what the
record showed, an optional successful contrast, and an optional
evidence-backed counterfactual.

Default to ONE behavioral mode. Split only when modes require
materially different behavioral_target values AND each has a
membership predicate that the other mode's members would fail.

A mode is publishable only if its members span at least TWO
distinct task ids. Put unsupported findings in residue. Every
finding id must appear exactly once. Do not name a harness
module, patch surface, prompt edit, or task-specific answer.

FINDINGS
{findings}

OUTPUT (one JSON object)
{
  "modes": [{
    "label":"UPPER_SNAKE",
    "mechanism":"one cross-task observable mechanism",
    "membership_test":"predicate over one Analyst finding",
    "behavioral_target":"smallest supported alternative" | null,
    "members":["finding_id", "..."]
  }],
  "residue":["finding_id", "..."]
}
\end{Verbatim}
\end{prompttemplate}

\begin{prompttemplate}{Harness Engineer \;|\; Evidence-linked source update}
\begin{Verbatim}[fontsize=\scriptsize]
You are the GUI-HARVEST Harness Engineer. Follow
ENGINEER_BRIEF.md exactly. Inspect only the baseline, candidate,
and frozen round evidence. Do not run GUI evaluation or decide
promotion.

YOUR JOB
1. Select one candidate, or a small compatible set explained by
   one shared mechanism.
2. Before editing, write proposal.pre.json and freeze targets,
   source attribution, predictions, invariants, and risks.
3. Edit only capability-approved candidate files; update source
   and HARNESS_CARD.md in the same edit.
4. Run local L0/L1 tests and write ENGINEER_REPORT.md.

proposal.pre.json  [schema: gui-harvest-intervention-v1]
{
  "proposal_id":"...",
  "targets":[{"candidate_id":"...", "lane":"...",
              "outcome_category":"..."}],
  "attribution":{"surface":"...", "mechanism":"...",
    "source_locations":[{"path":"...", "symbol":"...",
                         "anchor":"..."}]},
  "intervention":{"change_kind":"...", "summary":"...",
                  "genericity":"..."},
  "predictions":[{"prediction_id":"...",
    "target_candidate_ids":["..."], "statement":"...",
    "observable_test":"...", "target_task_ids":["..."],
    "required_for_promotion":true}],
  "invariants":["..."], "expected_regression_risks":["..."],
  "prior_attempts_addressed":[...]
}
\end{Verbatim}
\end{prompttemplate}

\begin{prompttemplate}{Behavioral Validator \;|\; Prediction--execution check}
\begin{Verbatim}[fontsize=\scriptsize]
You are the behavioral Validator in a GUI-agent harness experiment.
Judge only whether the frozen behavioral prediction is supported by
the post-edit trajectories. Do not judge task success, utility, or
promotion. Evaluate all supplied runs.

VERDICT RULES
- supported: relevant trigger(s) occurred and the claimed behavior held.
- contradicted: a relevant trigger occurred and a clear response
  violated the prediction.
- inconclusive: no relevant trigger occurred, evidence is missing,
  or the observable test cannot be decided from supplied rows.
- Never infer behavior from evaluator score; no score is supplied.
- Cite exact run labels and RESPONSE ROW step numbers.

OUTPUT (exactly one JSON object)
{
  "prediction_id":"...", "task_id":"...",
  "verdict":"supported|contradicted|inconclusive",
  "summary":"brief evidence-based explanation",
  "run_assessments":[
    {"run":"A", "verdict":"supported|contradicted|inconclusive",
     "reason":"..."}
  ],
  "citations":[
    {"run":"A", "step":3, "claim":"what this row proves"}
  ]
}
\end{Verbatim}
\end{prompttemplate}

\end{document}